%% file: main.tex
\documentclass[10pt,twocolumn,letterpaper,table]{article}

\usepackage[pagenumbers]{cvpr} %

\input{preamble}

\definecolor{cvprblue}{rgb}{0.21,0.49,0.74}
\usepackage[pagebackref,breaklinks,colorlinks,allcolors=cvprblue]{hyperref}

\def\paperID{6806} %
\def\confName{CVPR}
\def\confYear{2025}

\title{Light3R-SfM: Towards Feed-forward Structure-from-Motion}

\author{Sven Elflein$^{1,2,3}$ \quad Qunjie Zhou$^{1}$ \quad S{\'e}rgio Agostinho$^{1}$ \quad Laura Leal-Taix{\'e}$^{1}$ \vspace{0.3em} \\
{\normalsize $^1$NVIDIA} \quad
{\normalsize $^2$Vector Institute} \quad
{\normalsize $^3$University of Toronto}
}

\begin{document}
\maketitle

\input{sec/0_abstract}    
\input{sec/1_introduction}

\input{sec/2_related_work}
\input{sec/3_method}
\input{sec/4_experiment}

\input{sec/5_conclusion}

{
    \small
    \bibliographystyle{ieeenat_fullname}
    \bibliography{main}
}

\clearpage
\appendix
\twocolumn[{%
\begin{center}
\textbf{\Large Supplementary Material for \method}
\end{center}
}]

\input{sec/6_supplementary}

\end{document}

%% file: preamble.tex
\usepackage{graphicx}
\usepackage{amsmath}
\usepackage{amssymb}
\usepackage{booktabs}
\usepackage{multirow, makecell}
\usepackage{rotating}
\usepackage[normalem]{ulem}
\useunder{\uline}{\ul}{}
\usepackage{epigraph} 
\usepackage{dsfont}
\usepackage{bm}
\usepackage{enumitem} %
\usepackage[accsupp]{axessibility}
\usepackage{mathtools}

\usepackage{array}
\usepackage{arydshln}
\usepackage{hhline}
\usepackage[most]{tcolorbox} 
\usepackage[export]{adjustbox}
\usepackage{nicefrac}

\usepackage{pifont}%
\newcommand{\cmark}{\ding{51}}%
\newcommand{\xmark}{\ding{55}}%

\definecolor{cvprblue}{rgb}{0.21,0.49,0.74}
\newcommand{\PAR}[1]{\noindent{\bf #1~}}

\newcommand{\verttext}[1]{\rotatebox[origin=c]{90}{{#1}}}

\DeclarePairedDelimiter\floor{\lfloor}{\rfloor}

\newcommand{\method}{Light3R-SfM\xspace}
\newcommand{\mastr}{MASt3R\xspace}
\newcommand{\dustr}{DUSt3R\xspace}
\newcommand{\monstr}{MonST3R\xspace}
\newcommand{\spannr}{Spann3R\xspace}

\newcommand{\espt}{E_{\mathtt{SPT}}} %
\newcommand{\loss}[1]{\mathcal{L}_{\mathtt{#1}}}

\newcommand{\C}[2]{C^{#1,#2}} %
\newcommand{\X}[2]{X^{#1,#2}} %
\newcommand{\D}{\mathcal{D}} %
\newcommand{\bgcolor}[2]{\setlength{\fboxsep}{0pt}\colorbox{#1}{\strut #2}}

\definecolor{cbad}{HTML}{EAC2C2}
\definecolor{cmeh}{HTML}{FFE1C9}
\definecolor{cok}{HTML}{FDF3D0}
\definecolor{cgood}{HTML}{B3D09F}

\pgfdeclarehorizontalshading{rankgrad}{100bp}{
    color(0bp)=(cbad);
    color(20bp)=(cbad);
    color(40bp)=(cmeh);
    color(60bp)=(cok);
    color(80bp)=(cgood);
    color(100bp)=(cgood)
}

\newcommand{\BGcolor}[3][HTML]{\definecolor{mycolor}{HTML}{#2}\bgcolor{mycolor}{#3}}

\usepackage{pifont}%

%% file: sec/0_abstract.tex
\begin{abstract}

We present \method, a feed-forward, end-to-end learnable framework for efficient large-scale Structure-from-Motion (SfM) from unconstrained image collections. Unlike existing SfM solutions that rely on costly matching and global optimization to achieve accurate 3D reconstructions, \method addresses this limitation through a novel latent global alignment module. This module replaces traditional global optimization with a learnable attention mechanism, effectively capturing multi-view constraints across images for robust and precise camera pose estimation. \method constructs a sparse scene graph via retrieval-score-guided shortest path tree to dramatically reduce memory usage and computational overhead compared to the naive approach. Extensive experiments demonstrate that \method achieves competitive accuracy while significantly reducing runtime, making it ideal for 3D reconstruction tasks in real-world applications with a runtime constraint. This work pioneers a data-driven, feed-forward SfM approach, paving the way toward scalable, accurate, and efficient 3D reconstruction in the wild.

\end{abstract}

%% file: sec/1_introduction.tex
\section{Introduction}
\label{sec:intro}

\begin{figure}[t!]
    \centering
    \includegraphics[width=\linewidth]{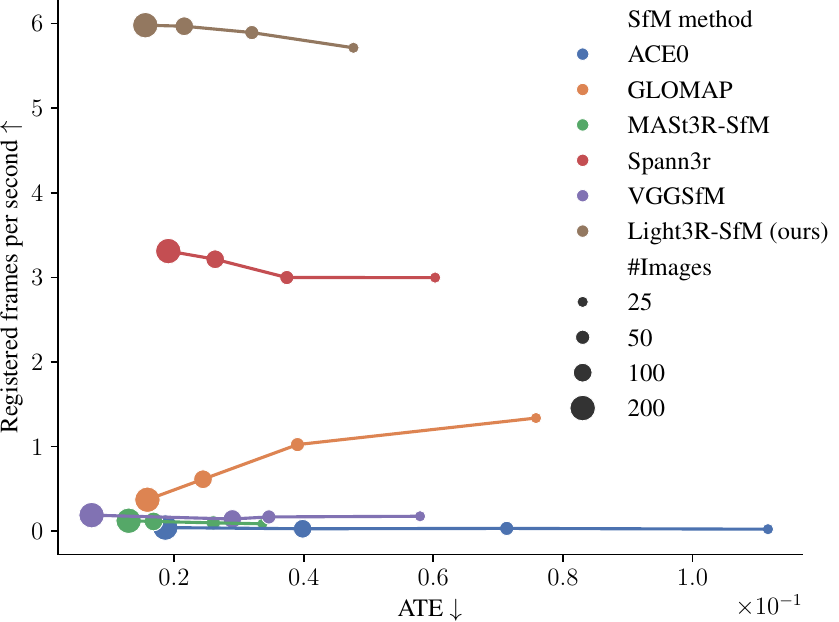}
    \caption{\textbf{Processing speed vs. accuracy for various SfM methods.} Our work significantly decreases the runtime across various sizes of image collections compared to traditional pipelines while obtaining comparable accuracy. Results are measured on the Tanks\&Temples dataset.}
    \label{fig:teaser}
    \vspace{-0.5cm}
\end{figure}

Structure-from-Motion (SfM) is the task of jointly recovering camera poses and reconstructing the 3D scene structure from a set of unconstrained images. This long-standing problem is essential to many computer vision applications, including novel view synthesis via NeRFs~\cite{mildenhall2021nerf, barron2021mipnerf} and 3DGS~\cite{kerbl20233dgs}, multi-view stereo (MVS) reconstruction~\cite{peng2022mvs, wang2023posediffusion}, and visual localization~\cite{sarlin2019hfnet, sattler2012activeserach}.
Traditional SfM methods generally follow two main approaches: incremental~\cite{snavely2006sfm, wu2013sfm, schonberger2016colmap} and global~\cite{pan2024glomap, wilson2011_1dsfm, cai2021globalsfm} SfM. Both paradigms rely on key components such as feature detection and matching for correspondence search, 3D triangulation to reconstruct geometry from 2D correspondences, and joint optimization of camera poses and scene geometry through bundle adjustment. A major research direction has been to replace these components with learning-based modules, progressing towards fully end-to-end SfM~\cite{brachmann2024acezero, smith2024flowmap, wang2024vggsfm}. 
\\
Recently, the seminal work~\dustr~\cite{wang2024dust3r} proposed to train an unconstrained stereo 3D reconstruction model through pointmap regression, \ie, by directly predicting 3D points in a common reference system for every pixel. Learning from large-scale annotated data, it shows impressive performance in handling
images with extreme viewpoint changes.
To perform SfM from an image collection, \dustr works~\cite{wang2024dust3r,leroy2024mast3r} first compute stereo reconstruction exhaustively for all image pairs and then obtain globally aligned pointmaps for all cameras through joint optimization of pairwise rigid transformations and local pointmaps. 
This baseline has been significantly improved by the concurrent work \mastr-SfM~\cite{duisterhof2024mast3rsfm} that leverages image retrieval to drastically reduce the computation overhead, boosts optimization efficiency by optimizing only over the sparse pixel correspondences, and appends a global bundle adjustment stage for accuracy refinement.
While optimization-based alignment has been proven to be the key to accurate 3D reconstruction by \dustr, \mastr-SfM and classical SfM methods~\cite{schonberger2016colmap, pan2024glomap, lindenberger2021pixsfm}, this comes at the cost of slow runtime and extensive memory footprint even for moderately-sized image collections.%

To this end, we propose \method, a fully learnable feed-forward SfM model that directly obtains globally aligned camera poses from an unordered image collection, without expensive optimization-based global alignment.
Instead, we perform implicit global alignment in the latent space with a scalable attention module between the image encoding and 3D decoding stages, which enables global information sharing between features before solving the pairwise 3D reconstruction.
This enables exploiting multi-view information across images, which is crucial for learning globally consist pointmaps.
\\
Concurrent work~\spannr~\cite{wang2024spann3r} tackles online reconstruction from videos by directly regressing pointmaps in a global coordinate system leveraging an explicit memory bank to store information from all previous frames and the current frame. %
The price paid for being an online model is that the memory bank is constrained by its fixed capacity and prone to drifting due to the propagation of errors over time. %
In contrast, our work focuses on offline reconstruction from unordered image sets. 
We exploit multi-view constraints via latent attention while minimizing the redundant processing supported by intelligent graph construction, delivering significantly more accurate camera poses with lower runtime than \spannr.
\\
We summarize the key contributions of this work as follows:
(i) We propose \method, a novel feed-forward SfM approach that replaces classical global optimization with a learnable latent alignment module, leveraging a scalable attention mechanism.
(ii) Through extensive experiments, we demonstrate that \method achieves more accurate globally aligned camera poses compared to the concurrent \spannr method. Its performance rivals state-of-the-art optimization-based SfM techniques while offering significant improvements in efficiency and scalability. Specifically, \method reconstructs a scene of 200 images in just 33 seconds, whereas the comparable \mastr-SfM takes approximately 27 minutes, resulting in a \textgreater\textbf{49 $\times$} speedup. We highlight the potential of fully feed-forward SfM and aim to inspire future research toward developing more reliable and accurate feed-forward methods for large-scale 3D reconstruction in real-world settings.

%% file: sec/2_related_work.tex
\section{Related Work}
\PAR{Classical SfM.}
Conventional structure from motion (SfM) methods can be divided into two main categories: incremental and global SfM.   Incremental SfM~\cite{snavely2006sfm, frahm2010sfm, agarwal2011sfm, wu2013sfm, schonberger2016colmap} approaches gradually reconstruct a 3D scene from  a collection of images starting from a carefully selected two-view initialization. Its main building blocks involve correspondence searching via feature detection and matching, pairwise pose estimation and 3D triangulation followed by bundle adjustment. 
Compared to incremental SfM, global SfM methods~\cite{pan2024glomap, wilson2011_1dsfm, cai2021globalsfm, cui2015globalsfm, arie2012globalsfm} start with a similar correspondence search and pairwise pose estimation stage, but then jointly align all cameras through rotation and translation averaging followed by 3D triangulation and bundle adjustment.
Global methods usually have faster runtime yet being less accurate and robust. 
A recent work GLOMAP~\cite{pan2024glomap} reduces its accuracy gap to incremental methods by combining the estimation of camera positions and 3D structure in a single global positioning step.
Our method, similar to the hybrid SfM method~\cite{liu2023hybridsfm}, divide images into subsets and incremental reconstruct the whole scene by accumulating locally aligned subsets.
However, we fundamentally differ from them by estimating camera poses for each local subset in a feed-forward manner through our learned deep network without requiring a further step of global bundle adjustment.
\\
\PAR{Optimization-based deep SfM.}
Recently, deep learning has been leveraged to improve core building blocks in SfM pipelines such as sparse feature detection~\cite{detone2018superpoint, revaud2019r2d2} and matching~\cite{sarlin2020superglue, lindenberger2023lightglue}. 
DFSfM~\cite{he2024dfsfm} adapts traditional keypoint-based SfM for leveraging dense feature matchers~\cite{zhou2021patch2pix, sun2021loftr, edstedt2024roma, edstedt2023dkm}. 
PixSfM~\cite{lindenberger2021pixsfm} introduces features-metric alignment to keypoints refinement and bundle adjustment, leading to improved accuracy and robustness under challenging conditions.
VGGSfM~\cite{wang2024vggsfm} adapts individual SfM components to their learned version forming a fully differentiable SfM framework.
Instead of optimizing camera parameters and scene geometries,
ACEZero~\cite{brachmann2024acezero} proposes a new learning-based SfM pipeline to incrementally optimizing scene coordinate regression and camera refinement networks that output camera parameters and geometries through reprojection errors. 
Similarly, FlowMap~\cite{smith2024flowmap} presents another end-to-end differentiable SfM pipeline where a depth estimation network is optimized per-scene through offline optical flow and point tracking supervisions.
Yet, those methods~\cite{brachmann2024acezero, smith2024flowmap} struggle with image pairs with low visual overlapping.
\\
Recently, the emerging unconstrained stereo 3D reconstruction model, \dustr~\cite{wang2024dust3r}, opened up a new paradigm for tackling 3D reconstruction tasks such as SfM and multi-view stereo (MVS). Essentially, it proposed a radically novel approach for a two-view reconstruction via direct pointmap regression from a pair of RGB images. Different from monocular scene coordinate regression~\cite{brachmann2017dsac, brachmann2023ace, shotton2013sevenscene, brachmann2024acezero}, such stereo pointmap regression formulation can benefit from large-scale training achieving strong generalization capability to new scenes.
To perform SfM from an image collection, \dustr~\cite{wang2024dust3r} and its improved version \mastr~\cite{leroy2024mast3r} merge pairwise pointmap predictions via optimization-based global alignment, however, exhaustive pairwise pointmaps in a brute-force manner limits their application to a small set of images~\cite{brachmann2024acezero}. 
To tackle this, \mastr-SfM~\cite{duisterhof2024mast3rsfm} incorporates image retrieval exploiting the \mastr encoder feature embedding to build a sparse scene graph, leading to significantly reduced runtime. Additionally, it leverages sparse correspondences to boost optimization efficiency and accuracy.
Another concurrent work \monstr~\cite{zhang2024monst3r} extends \mastr to handle dynamic scenes by leveraging an offline optical flow estimation.
However, those \dustr-based SfM methods still rely on expensive iterative optimization for accurate globally aligned poses. 

\PAR{Feed-forward SfM.}
Instead of performing optimization-based global alignment, \spannr~\cite{wang2024spann3r}, 
leverages explicit spatial memory to implicitly align pointmaps \wrt the first frame, which requires to maintain spatial information for all following frames in a sequence of arbitrary length overtime.
Compared to \spannr, our proposed feed-forward SfM efficiently exploits multi-view constraints from a collection of unconstrained images at the same time, utilizing a scalable latent alignment module together with efficient graph construction, leading to more efficient and accurate global alignment of pointmaps than \spannr.

%% file: sec/3_method.tex
\section{\method}
\label{sec:method}

\begin{figure*}[htp!]
    \centering
    \includegraphics[width=0.85\linewidth]{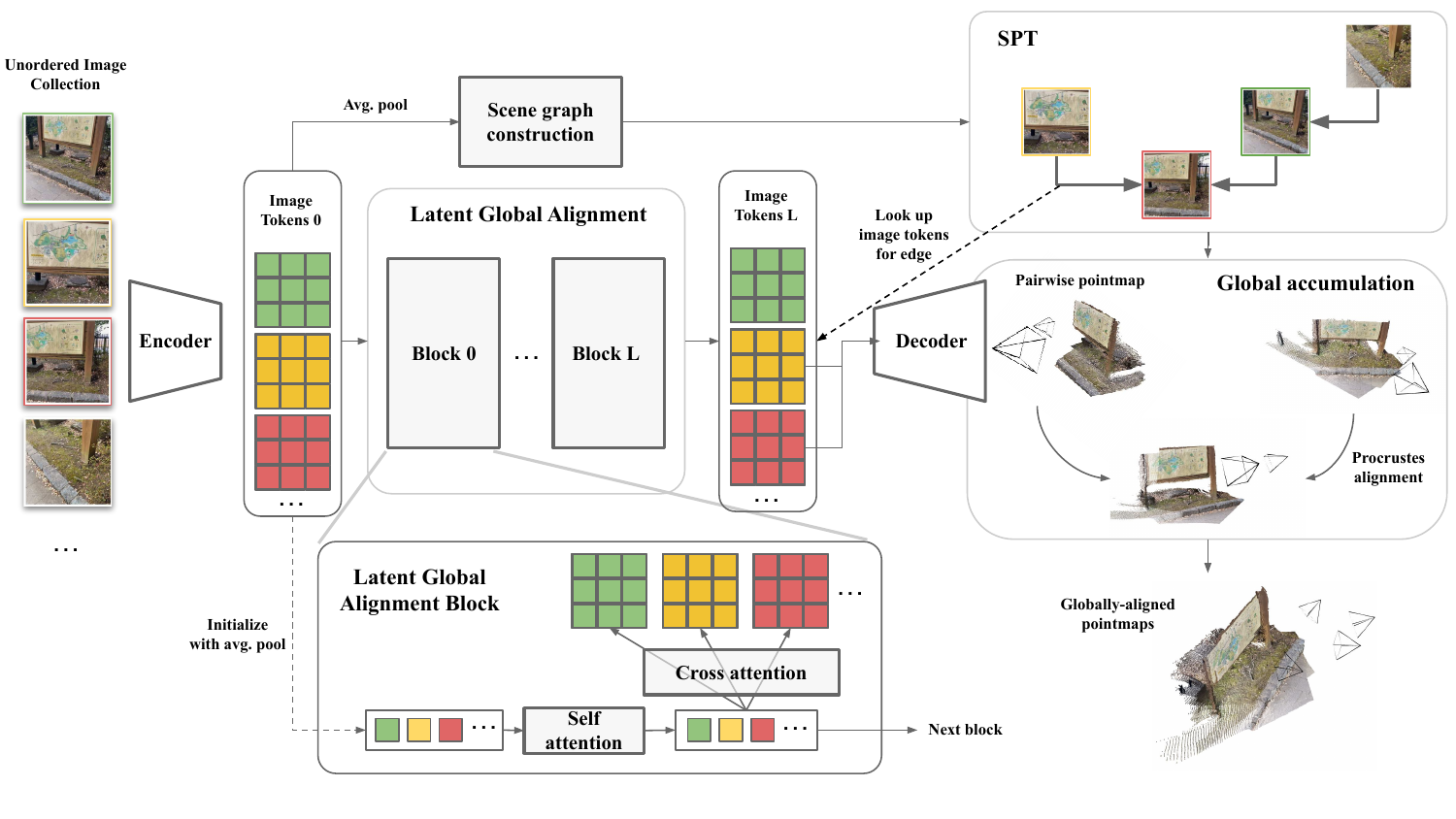}
    \vspace{-7mm}
    \caption{\textbf{\method Pipeline.} Given an unordered set of images,  we first encode them to obtain image tokens from which we average pool global features for constructing a shortest path tree. We next feed image tokens into our attention-based latent global alignment to enable global context sharing. Afterwards, for each edge in the SPT, we decode pairwise pointmaps using the implicitly aligned feature tokens. Finally, we use global accumulation to obtain globally aligned pointmaps per image.}
    \vspace{-0.2cm}
    \label{fig:pipeline_overview}
\end{figure*}

In this section, we present \method, a novel feed-forward SfM model that enables robust, accurate and efficient structure-from-motion in the wild for large-scale real-world applications. The key component is an attention mechanism that allows optimization-free globally aligned pose estimation for the entire image set.

Given an unordered image collection or a sequence of images, denoted as $\{\mathcal{I}_i\}_{i=1}^{N}$ with $\mathcal{I}_i \in \mathbb{R}^{H \times W \times 3}$, our pipeline reconstructs per image camera extrinsics $P\in \mathbb{R}^{4 \times 4}$, intrinsics $K_i \in \mathbb{R}^{3\times3}$ and dense 3D pointmap at image resolution $X \in \mathbb{R}^{H \times W \times 3}$, which represents the globally aligned scene geometry observed by individual images.
As shown in \cref{fig:pipeline_overview}, we start with the (i) \textit{encoding}, where an image encoder extract per-image feature tokens. After that we have the (ii) \textit{latent global alignment}, in which information is exchanged between all image tokens via a scalable attention mechanism to globally align image tokens in the feature space (\cref{method:global_alignment}). Next, the (iii) \textit{scene graph construction} constructs a scene graph maximizing pairwise image similarities via running the shortest path tree (SPT) algorithm. The (iv) \textit{decoding} step converts image pairs connected by an edge to pointmaps using a stereo reconstruction decoder (\cref{method:graph_construction}). Finally, we run the (iiv) \textit{global optimization-free reconstruction}, which accumulates the pairwise pointmaps by traversing the scene graph (\cref{method:accumulation}) to obtain the globally aligned pointmaps.

\subsection{Latent Global Alignment}
\label{method:global_alignment}
We start by encoding each image $\mathcal{I}_i$ to image tokens
\begin{equation}
F^{(0)}_i = \mathtt{Enc}(\mathcal{I}_i), \quad F^{(0)}_i \in \mathbb{R}^{\floor{H / p} \times \floor{W / p} \times d}, 
\end{equation}
where $p$ is the patch size of the encoder and $d$ is the token dimensionality. 
To allow information sharing between all images without running into memory constraints, we take inspiration from ~\citet{karaevCoTrackerItBetter2023}, who apply a similar principle to point tracks, and factorize the attention operation between all frames via a smaller set of tokens.
Specifically, for each set of image tokens $F^{(0)}_i$ we compute its global token $g_i^{(0)} \in R^{d}$ via averaging along its spatial dimensions.
We then use $L$ blocks of our latent global alignment block to achieve global information sharing across all image tokens.
For each level $l \in (0, L)$, we first share information across all global image tokens $\{g^{(l)}_i\}_{i=1}^{N}$ using self-attention defined as
\begin{equation}
    \{g_i^{(l+1)}\}_{i=1}^{N} = \mathtt{Self}(\{g_i^{(l)}\}_{i=1}^{N}).
\end{equation}
We then propagate the updated global information to dense image tokens $\{F^{(l)}_i\}_{i=1}^{N}$ for each image independently via cross-attention:
\begin{equation}
    F_i^{(l+1)} = \mathtt{Cross}(F_i^{(l)}, \{g_i^{(l+1)}\}_{i=1}^{N}).
\end{equation}
Finally, we obtain the globally aligned image tokens $F_i$ via a residual connection, \ie,  $F_i \coloneq F_i^{(0)} + F_i^{(L)}$.

\PAR{Discussion.}
A naive implementation through self-attention between all image tokens requiring $\mathcal{O}((N \times T)^2)$, while our latent global alignment module is able to achieve a time complexity of $\mathcal{O}(N^2 + N \times T)$, where $T=\floor{H / p} \times \floor{W / p}$ is the number of per-image tokens and $N$ is the number of images. 
While the same asymptotic complexity class, we find reducing the constant factor for practical values of $N\approx T$ to be the key to scale to larger image collections.

\subsection{Scene Graph Construction}
\label{method:graph_construction}
Despite our global feature attention being lightweight by design, exhaustively decoding 3D pointmaps for all image pairs through a fully connected scene graph still leads to a computational bottleneck.
We thus propose a more scalable approach to scene graph construction allowing us to decode pointmaps with just $N - 1$ edges.
For that, we leverage the encoder embeddings to compute pairwise similarities similar to concurrent work~\cite{leroy2024mast3r}, which allows us
to filter out irrelevant image pairs, \eg, pairs with low visual overlap, to avoid unnecessary computation~\cite{duisterhof2024mast3rsfm, yan2021irsfm}.
Specifically, we average pool the tokens of each image $F_i$  to obtain one-dimensional embedding $\bar{F}_i$ and then compute the matrix $S$ containing all pairwise cosine similarities as
\begin{equation}
S_{ij} = \langle \| \bar{F}_i \|_2,   \| \bar{F}_j \|_2 \rangle
\end{equation}
where $\langle \cdot, \cdot \rangle$ denotes the scalar product.
Classical SfM methods~\cite{schonberger2016colmap,pan2024glomap} build a scene graph as a minimum spanning tree (MST) which 
minimizes the sum of costs, \ie, the negative similarities, of \textit{all} edges.
However, this often results in trees with high depth that
result in drift when we accumulate pairwise pointmaps, as proposed in the next section.

Therefore, we propose to replace the MST with a shortest path tree (SPT)~\cite{ford_spt} to obtain a scene graph as a set of edges $\espt = \{(i, j)\}$ connecting all images, while minimizing the cost of the paths towards each node. Intuitively, this leads to a flatter tree which only runs deep when it benefits the overall reconstruction.
We set the root node for the SPT as the one with lowest total cost \wrt all other nodes, \ie, $\operatorname*{argmin}_{j}  \sum_i -S_{ij}$.  
By design, the number of edges in a tree is linear in the number of images $N$, \ie, $\left| \espt \right| = N - 1$, leading to significantly better scalability than a fully-connected graph.

\subsection{Global Optimization-free Reconstruction}
\label{method:accumulation}
To obtain the global reconstruction while still being end-to-end trainable, we first obtain per-edge local pointmap predictions and then merge local pointmaps into a global one.

\PAR{Edge-wise pointmap decoding.} For every edge in the scene graph $(i, j) \in \espt$, we run the decoder to output two pointmaps and associated confidence maps defined as:
\begin{equation}
  (\X{i}{i}, \X{j}{i}), (\C{i}{i}, \C{j}{i}) = \mathtt{Dec}\left(F_i, F_j\right).
\label{eq:dec}
\end{equation}
Here, $\X{i}{i} \in \mathbb{R}^{H \times W \times 3}$ is the pointmap of the $i$-th image and $\X{j}{i}$ is the pointmap of the $j$-th image, both in the coordinate frame of the $i$-th image. 
$\C{j}{i}, \C{i}{i} \in \mathbb{R}^{H \times W}$ are the per-point confidence scores for each pointmap respectively.
While this closely follows the setup in \cite{wang2024dust3r}, note that the input features to the decoder are conditioned on all images which facilitates globally aligned pairwise pointmaps.

\PAR{Global accumulation.} 
To combine the pairwise pointmap predictions into a global reconstruction $\mathbf{X}$ with per-point confidences $\mathbf{C}$, we traverse the SPT $\espt$ in breadth-first order, starting from the root of the tree. 
For the first edge, we initialize the global point cloud as $\mathbf{X} = \{X^{i}, X^{j}\}$ and $\mathbf{C} = \{C^{i}, C^{j}\}$ where $X^i \coloneq \X{i}{i}$ and $X^{j} \coloneq \X{j}{i}$ are the pointmap predictions for the edge in the coordinate system of the $i$-th image and $C^{i}, C^{j}$ their corresponding confidences.
The $i$-th camera is thus implicitly defined as the canonical frame for the global reconstruction.
We next register the remaining local reconstructions predicted from the consecutive edges to this initial global reconstruction.

Based on the traversal order, node $k$ of the next edge $(k, l)$ has already obtained its global registered pointmap $X^{k} \in \mathbf{X}$ in the previous step.
We first update the global confidence map of this node to $C^k \coloneq C^k \odot \C{k}{k}$, where $\odot$ denotes the element-wise geometric mean, to take into consideration the confidence of the pointmap prediction $\C{k}{k}$ given the current pair.
To register the $l$-th node to the global reconstruction, we then estimate the  optimial rigid body transformation between the two pointmaps, $X^{k}$  (in the global coordinate) and  $\X{k}{k}$ (in the same coordinate system of the $l$-th node) via Procrustes alignment~\cite{umeyamaLeastsquaresEstimationTransformation1991}
\begin{equation}
P_{k} = \mathtt{Procrustes}(X^{k}, \X{k}{k}, \log C^{k})
\end{equation}
where $\log C^{k} \in \left[0, \infty\right]^{H \times W}$ serves as a per-point weight.
Finally, we transform the pointmap of node $l$ into the global coordinate frame
\begin{equation}
\X{l}{} = P_k^{-1} \X{k}{l}
\end{equation}
and add it the global reconstruction $\mathbf{X} \coloneq \mathbf{X} \cup \{X^{l}\}$.
Repeat it for all edges in $\espt$, we obtain per-image globally registered pointmaps $X^{i}$ with associated 
confidences $C^{i}$ .

\PAR{Discussion.} 
While our method still involves Procrustes alignment for each node, it is a significantly simpler problem compared to jointly optimizing a large number of 3D points and camera parameters among all images, which is much more sensitive to noisy pointmap and confidence predictions and limited to small number of images. 
Furthermore, compared to iterative solvers used in bundle adjustment, Procrustes alignment can be efficiently solved in closed form, and thus its computation overhead, linear in the number of images, is negligible.

\subsection{Supervision}
\label{method:supervision}
We jointly supervise pairwise local pointmaps and the globally aligned pointmaps, with the focus of the latter to enforce accurate and consistent global alignment learning.

\PAR{Pairwise supervision.}
Given a set of ground-truth pointmaps in the world coordinate frame $\mathbf{\bar{X}} = \{\bar{X}^i\}_{i=1}^{N}$, the corresponding valid pixels $\{\D^i\}_{i=1}^{N}$, and the ground truth camera poses ${P^i}_{i=1}^{N}$,
we compute $\loss{pair}$ that supervises the pairwise local pointmaps per-edge in the coordinate frame of the first camera following \dustr~\cite{wang2024dust3r}:
\begin{multline}
  \loss{pair} = \sum_{(i, j) \in \espt} (\loss{conf}(P_i \bar{X}^i, \X{i}{i}, \C{i}{i}, \D^i) \\
  + \loss{conf}(P_i \bar{X}^j, \X{j}{i}, \C{j}{i}, \D^j) ) \,, 
\label{eq:pairwise_supervision}
\end{multline}
\begin{equation}
\loss{conf}(\bar{X}, X, C, \D) \coloneq \sum_{p \in \D} C_p \left\Vert X_{p}  - \bar{X}_p \right\Vert - \alpha C_p.
\label{eq:regression}
\end{equation}
Here $X, C, \bar{X}$ are the predicted pointmap, confidence map and the ground-truth pointmap, $D \subseteq \{1\ldots W\}\times\{1\ldots H\}$ defines the valid pixels with ground-truth, and $\alpha > 0$ regularizes the confidences to not be pushed to $0$~\cite{wanConfnetPredictConfidence2018}.

\PAR{Global supervision.}
We first align the global pointmaps $\mathbf{X} = \left\{\X{1}{}, \ldots, \X{N}{}\right\}$ that are defined \wrt the root node of the SPT, to the ground truth pointmaps, by estimating the optimal rigid body transformation:
\begin{equation}
\smash{
    P_{\mathtt{align}} = \mathtt{Procrustes}(\mathbf{\bar{X}}, \mathbf{X}).
}
\end{equation}
\\
We then supervise the transformed global pointmap prediction for each image as
\begin{multline}
  \loss{global} = \sum_{i \in \{1, \ldots, N\}} \loss{conf}(\bar{X}^i, P_{\mathtt{align}} X^i, C^i, \D^i)
\label{eq:supervision_global}
\end{multline}
In practice, we do not compute the loss for samples with less than $100$ valid pixels due to inaccurately estimated rigid body transformation.
This loss implicitly supervises the accuracy of poses extracted from these pointmaps since 
inaccurate poses from the pairwise Procrustes alignment 
leads to higher global loss.
We optimize $\mathcal{L} = \loss{pair} + \lambda \loss{global}$, empirically setting $\lambda = 0.1$.

%% file: sec/4_experiment.tex
\section{Experiments}
\label{sec:experiments}
In this section, we conduct extensive evaluation across diverse datasets and scenes, covering a wide range of typical SfM settings, and thorough ablations to understand our model. We provide implementation details for training and inference in the supplementary material.

\subsection{Scene-level Multi-view Pose Estimation}\label{subsec:exp_t&t}
\PAR{Dataset.} 
We first evaluate our method on multi-view pose estimation using Tanks\&Temples~\cite{Knapitsch2017TanksAndTemples} covering 21 indoor and outdoor scenes, where each scene contains 150-1100 images with uncalibrated cameras~\cite{brachmann2024acezero}.

\PAR{Baselines.}
We categorize our SfM baselines into two main categories, \ie, \textit{optimization-based} (OPT) and \textit{feedforward-based} (FFD) methods, according to their global alignment methodology. 
For \textit{optimization-based} methods, we consider the classical SfM pipelines Colmap~\cite{schonberger2016colmap} (with SuperPoint~\cite{detone2018superpoint} and SuperGlue~\cite{sarlin2020superglue}), DF-SfM~\cite{he2024dfsfm}, Glomap~\cite{pan2024glomap}, PixelSfM~\cite{lindenberger2021pixsfm} and VGGSfM~\cite{wang2024vggsfm}, the end-to-end SfM including ACE-Zero~\cite{brachmann2024acezero} and FlowMap~\cite{smith2024flowmap}, as well as the recent state-of-the-art \mastr-SfM~\cite{duisterhof2024mast3rsfm}.
For \textit{feedforward-based} methods, we compare to our only baseline, the concurrent method \spannr, where we evaluate both its online and offline version whenever possible.

\PAR{Metrics.}
Given a set of images, we follow previous work~\cite{wang2024dust3r, duisterhof2024mast3rsfm, pan2024glomap} to compute the relative camera pose errors for all image pairs and measure the percentage of pairs with angular rotation/translation error below a certain threshold~$\tau$, denoted as relative rotation accuracy (RRA$@\tau$) and relative translation accuracy (RTA$@\tau$). We report its accuracy score average over all data samples.
We further report the percentage of successfully registered images (Reg.) where we count failed scenes with a registration rate of 0, and average translation errors (ATE) where we align estimated camera positions to the ground-truth (estimated using Colmap using all frames provided and provided by \cite{duisterhof2024mast3rsfm}) with Procrustes~\cite{umeyamaLeastsquaresEstimationTransformation1991} and report an average normalized error. We also report the runtime for a subset of methods on a system with a NVIDIA V100-32GB.

\begin{table}[t!]
    \centering
    \resizebox{1.\linewidth}{!}{\input{tables/tt_overview_with_mast3r-sfm_values_colored}}    
    \vspace{-0.25cm}
    \caption{\textbf{Multi-view pose estimation on Tanks\&Temples~\cite{Knapitsch2017TanksAndTemples}.} We adopt the benchmark by \cite{duisterhof2024mast3rsfm} and consider 25/50/100/200 view subsets and using the full sequence. We report relative pose accuracy RRA@5 and RTA@5, absolute translation error (ATE) and registration rate (Reg.). For clarity, we color-code results with a linear gradient between the \BGcolor{e0a4a4}{ w}\BGcolor{e7b2ac}{o}\BGcolor{eec0b5}{r}\BGcolor{f6cebe}{s}\BGcolor{fcdcc6}{t}\BGcolor{ffe4ca}{ }\BGcolor{fee8cc}{a}\BGcolor{feeccd}{n}\BGcolor{fdf0cf}{d}\BGcolor{f6f0cb}{ }\BGcolor{e3e6bd}{b}\BGcolor{cfdcae}{e}\BGcolor{bbd2a0}{s}\BGcolor{a8c992}{t } result for a given scene. `-' results indicate that all scenes did not converge or that we did not obtain runtime measurements. We specify the type of alignment used by each methods, `OPT' stands for \textit{optimization-based} and `FFD' stands for \textit{feedforward-based}. 
}        
    \vspace{-0.4cm}
    \label{tab:tandt}
\end{table}

\PAR{Comparison to state-of-the-art methods.}
We follow previous work~\cite{duisterhof2024mast3rsfm} compare across 5 different view settings including sparsely sampled 25/50/100/200 frame subsets and the original full sequence.
As shown in \cref{tab:tandt}, our method is competitive with other learning-based methods including VGGSfM, ACE-Zero, and FlowMap. Our method is less accurate than Glomap, Colmap and the concurrent work \mastr-SfM, particularly for the dense view setting with more than 200 images.
Those methods rely on classical optimization techniques such as bundle adjustment~\cite{duisterhof2024mast3rsfm, schonberger2016colmap, pan2024glomap, he2024dfsfm} or 3D global alignment~\cite{wang2024dust3r, duisterhof2024mast3rsfm} to achieve better accuracy, but they suffer from limited scalability. For example, Glomap and \mastr-SfM require $30\times$ and $43 \times$ more runtime than our method in full-sequence.
\\
The only method that has a runtime in the same magnitude as ours is \spannr, which also proposes to replace the computationally expensive global alignment of \dustr with an implicit alignment implemented via a memory bank.
To enable \spannr online evaluation, we sort the multi-view image frames based on their timestamps.  
Compared to \spannr, we show that our latent global alignment module is significantly superior for the SfM setting in both accuracy and runtime, leading to an average of $145\%$ and $84\%$ increase in RRA and RTA scores across 5 view settings with approximately half of its runtime.

\begin{table}[t!]
    \centering
    \resizebox{\linewidth}{!}{\input{tables/ours_vs_spann3r_new}}
    \vspace{-0.25cm}
    \caption{\textbf{Detailed comparison to \spannr.} We compare on Tanks\&Temples using the 25 and 50 image subsets as well as the full sequences.}
    \vspace{-0.3cm}
    \label{tab:abl-vs-spannr}
\end{table}

\PAR{Detailed comparison to \spannr.} 
To fully demonstrate the advantage of our proposed approach for feed-forward SfM, we present a more thorough comparison to the other feed-forward method \spannr. 
As shown in \cref{tab:abl-vs-spannr}, \spannr degrades significantly if the input is an unordered image set due to the lack temporal coherence between frames.
To process an unordered image set, \spannr proposes an offline version that relies on estimating the optimal order of frames via exhaustively evaluating all images pairs. We find that this leads to a significant runtime increase, \eg, $\times 36$ higher runtime than our \method in the 50-view setting, while still being significantly less accurate. When moving to the full video sequences with up to 1100 frames, this setup even encounters out-of-memory errors.
Finally, even when operating on the full, sorted video sequence, which \spannr is specifically designed to handle, performance is still subpar. In comparison, our method is able to provide more accurate poses, even when considering the same image resolution, fully validating the superiority of our method for the SfM setting.
\subsection{Object-centric Multi-view Pose Estimation}\label{subsec:exp_co3d}
\PAR{Dataset.} Next, we evaluate our method on object-centric scenes, for which we consider CO3Dv2~\cite{reizensteinCommonObjects3D2021}, containing 37k turntable-like videos of objects from 51 MS-COCO categories and camera poses annotated using Colmap.

\PAR{Baselines.} In addition to the previously mentioned baselines, we further compare to the state-of-the-art multi-view pose regression methods including PoseDiff~\cite{wang2023posediffusion}, PosReg~\cite{wang2023posediffusion}, and RelPose++~\cite{lin2023relpose++}, which also predict aligned camera poses in a feed-forward manner. 
We also compare to a \mastr\textbf{*} baseline where we employ the off-the-shelf \mastr model without any additional training inside our proposed framework.
\PAR{Metrics.} We report relative rotation accuracy and relative translation accuracy with threshold 15, \eg, RRA@15 and RTA@15. 
We further calculate the mean Average Accuracy (mAA)@30, defined as the area under the curve accuracy of the angular differences at $\min$(RRA@30, RTA@30).
\begin{table}[t!]
    \centering
    \resizebox{\linewidth}{!}{\input{tables/co3d_10views}}
    \vspace{-0.55cm}
    \caption{\textbf{Wide-baseline, multi-view camera pose estimation on CO3Dv2~\cite{reizensteinCommonObjects3D2021}}. We vary the number of input images by randomly sampling from the original sequence.             
       }
    \vspace{-0.3cm}
    \label{tab:sparse_view_co3d_r10k}
\end{table}
\PAR{Results.}
Similar to \cite{duisterhof2024mast3rsfm}, we evaluate on 2-view and 10-view settings.
As shown in the \textit{upper} part of \cref{tab:sparse_view_co3d_r10k}, for the 10-view setting, we are significantly more accurate than traditional \textit{optimization-based} SfM methods such as Colmap, Glomap and PixSfM.
Our performance is on-par with VGGSfM which trains one model per dataset for evaluation, while our method generalizes across datasets.
We are less accurate than \dustr and concurrent \mastr-SfM that benefit from \textit{optimization-based} global alignment at the cost of scalability and efficiency as discussed in \cref{subsec:exp_t&t}, however, we note that our latent global alignment almost closes the gap towards \dustr-GA.
Compared to other \textit{feed-forward} methods, our method achieves the best performance across all metrics.
Comparing to \mastr\textbf{*}, we demonstrate the benefit coming from our contributions leading to a $6.1\%$ increase in RTA@15.
Among the feed-forward methods, only ours and \spannr evaluate on a large number of images, while the other works typically focus on object-centric scenes.
In the \textit{bottom} part of \cref{tab:sparse_view_co3d_r10k}, we demonstrate superior performance compared to other \textit{feed-forward} methods on pairwise pose estimation. This setting decouples the need for global alignment, indicating that our proposed global supervision on the accumulated pairwise predictions helps to predict more accurate pointmaps for pose estimation in general.
\begin{table}[t!]
    \centering
    \resizebox{\linewidth}{!}{\input{tables/waymo}}
    \caption{\textbf{Camera pose estimation on driving scenes.}}
    \label{tab:waymo}
    \vspace{-0.3cm}
\end{table}
\subsection{Evaluation on Driving Scenes}
\PAR{Dataset.} We further evaluate our model on driving scenes where we validate our generalization capability to translating camera motion. We use the validation split of Waymo Open Dataset~\cite{waymo_dataset}, which contains a collection of 200 20-second clips recorded at 10Hz from an autonomous vehicle. 
For each sequence, we use the 200 input frames from the forward-looking camera.

\PAR{Baselines.} We compare against our concurrent works \spannr and \mastr-SfM, which also build on top of \dustr and have also seen Waymo training split as us. 

\begin{figure}[t!]
    \centering
    \includegraphics[width=\linewidth]{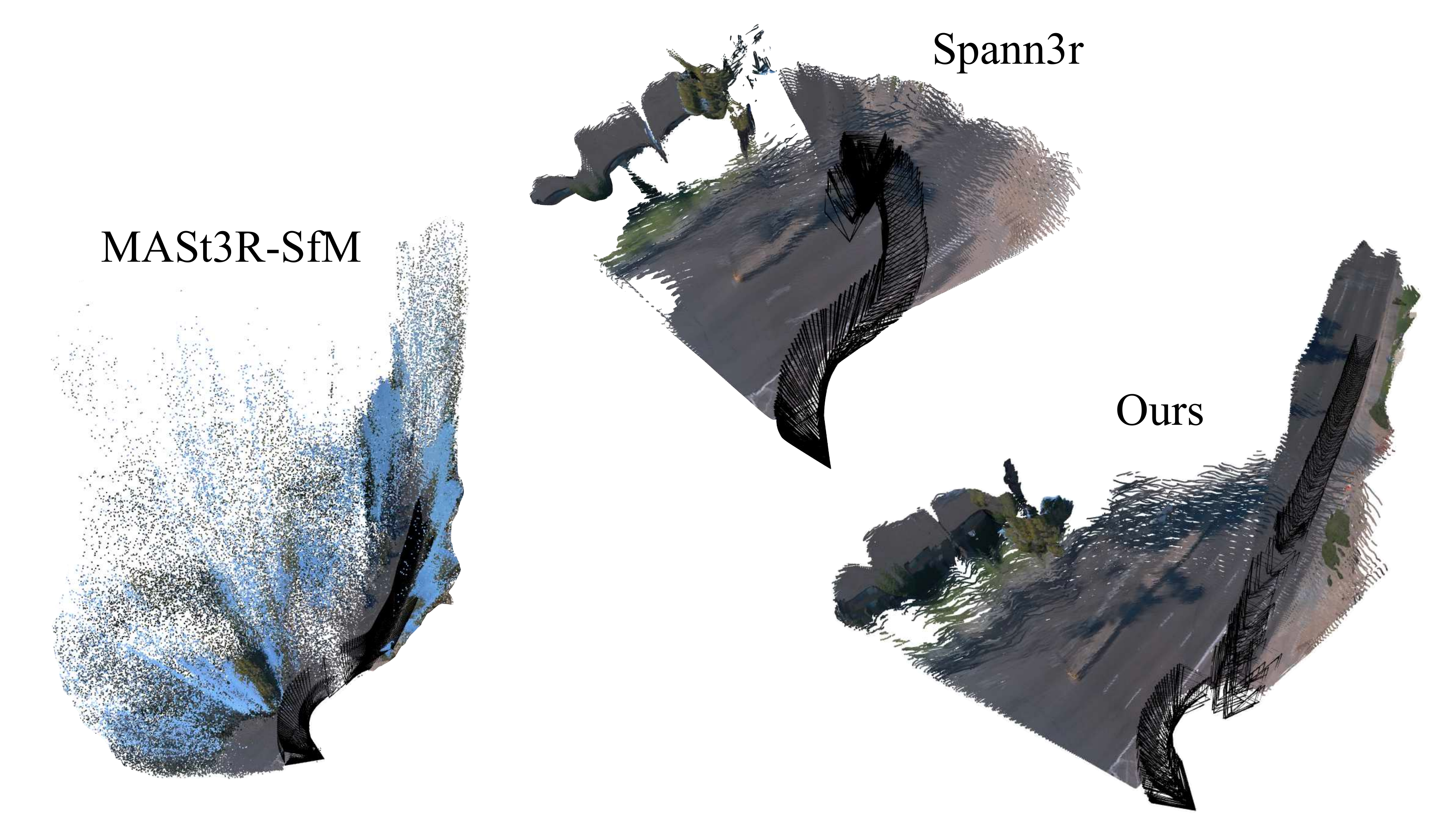}
    \vspace{-0.4cm}
    \caption{\textbf{Qualitative comparison on a Waymo scene.} Note how the \mastr-SfM reconstruction does not truthfully reconstruct the 90\textdegree{} turn, while \spannr predictions degrade after tens of frames.}
    \label{fig:waymo_comparison}
    \vspace{-0.2cm}
\end{figure}

\PAR{Results.} 
As shown in \cref{tab:waymo}, \method achieves comparable accuracy to \textit{optimization-based} \mastr-SfM at $\sim 195 \times $ lower runtime.
Compared to the concurrent \textit{feed-forward} \spannr, we significantly outperform \spannr achieving $\sim 4\times$ better accuracy in RTA@5 at \textgreater $6\times$ lower runtime.
This confirms that our latent alignment design leads to a more accurate and efficient modeling of global alignment compared to a memory-based architecture in \spannr.
We visualize pointmaps predicted by \method and our baselines in \cref{fig:waymo_comparison}.

\subsection{Ablation Studies}
In this section, we perform in-depth ablation study to fully understand each component of our method.  We perform all ablation experiments on Tanks\&Temple~\cite{Knapitsch2017TanksAndTemples} dataset using its 200 view subset and report the same multi-view pose estimation metrics as defined in ~\cref{subsec:exp_t&t}.

\begin{table}[t!]
    \centering
    \resizebox{\linewidth}{!}{\input{tables/ablat_model}}
    \vspace{-0.25cm}
    \caption{\textbf{Model ablation.} We study the impact of backbone initialization, global supervision, latent alignment as well as different ways for graph construction on pose estimation performance.
    }
    \vspace{-0.2cm}
    \label{tab:abl-model}
\end{table}

\PAR{Model components.}
In the part \textbf{(a)} of \cref{tab:abl-model}, we ablate the influence of backbone initialization, global supervision defined in \cref{eq:supervision_global} and latent alignment defined in \cref{method:global_alignment} when using the same graph construction process and pose accumulation process described in \cref{method:accumulation}. 
We show that adding latent alignment leads to 6.95\% increase in RRA@5 and 15.78\% decrease in ATE over the baseline. 
By further adding global supervision, we obtain our final model (\textit{last row}), which brings another 3.14\% and 9.03\% increase in RRA@5 and RTA@5.
Switching the backbone initialization of our full model from \mastr to \dustr leads to 6.87\% and 8.1\% drop in RRA@5 and RTA@5.

\PAR{Scene graph construction.}
In part \textbf{(b)} of \cref{tab:abl-model}, we analyze the impact of different options in building the scene graph introduced in \cref{method:graph_construction}.
Specifically, we present an graph construction Oracle, where we use ground truth overlapping score to obtain an optimal spanning tree, indicating the upper bound performance our method can achieve by improving the retrieval step. 
We show that our encoder retrieval is able to construct a high-quality scene graph that leads to performance very close to Oracle.
We further compare to another baseline where we obtain a minimal spanning tree instead. This leads to 15.26\% and 25.61\% accuracy decrease in RRA@5 and RTA@5, demonstrating the importance modulating the depth of the tree for our method.

\begin{table}[t!]
    \centering
    \resizebox{1.\linewidth}{!}{\input{tables/pointmap_conf_analysis}}    
    \vspace{-0.25cm}
    \caption{\textbf{Pointmap confidence analysis on Tanks\&Temples~\cite{Knapitsch2017TanksAndTemples}.} Our learned confidence maps effectively filter \textit{outlier} points, leading to increased rejected frames yet overall more accurate poses.
    }
    \vspace{-0.3cm}
    \label{tab:pointmap_conf}
\end{table}

\PAR{Pointmap confidence analysis.}
During inference, we compute global camera poses for each frame from the predicted pointmaps where we filter out 3D points if their confidence score is lower than a certain confidence threshold. 
In \cref{tab:pointmap_conf}, we study the impact of different confident thresholds, \eg, from 3 to 7, on pose estimation performance.
We show that the learned confidence maps effectively identify the confident points from images, leading to decreasing registration rate and improved pose accuracy when we increase the threshold, which enables a flexible control over trading-off accurateness and completeness depending on the downstream applications.

\PAR{Generalization.}
We showed that our method trained on a specific  8-view graph structure generalizes well to all kinds of view settings, \eg, (minimal) 2-view,  (sparse) 10/25/100/200-view and full-view settings in \cref{tab:tandt}.
In addition, we confirm solid generalization of our method by consistently outperforming our concurrent baseline \spannr, across different types scenes and datasets including object-centric scenes from Co3Dv2~\cite{reizensteinCommonObjects3D2021}, driving scenes from Waymo~\cite{waymo_dataset} and \textit{unseen} natural indoor and outdoor scenes from Tanks and Temples ~\cite{Knapitsch2017TanksAndTemples}.

\begin{table}[t!]
    \centering
    \resizebox{1\linewidth}{!}{\input{tables/runtime}}
    \caption{\textbf{Runtime Analysis.} We evaluate \method on the Courthouse scene with 1106 images using a NVIDIA V100-32GB.
    }
    
    \vspace{-0.1cm}
    \label{tab:abl-runtime}
\end{table}
\PAR{Runtime analysis.}
We analyze the detailed runtime required by individual components and the memory footprint of our method in \cref{tab:abl-runtime} by evaluating it on the Courthouse scene at two image resolutions using a NVIDIA V100-32GB. We split the batch into chunks of 32 for both the encoder and decoder which can be adapted to fit smaller or larger GPU memory budgets, trading off runtime.

\begin{figure}[tp!]
    \centering
    \includegraphics[width=\linewidth]{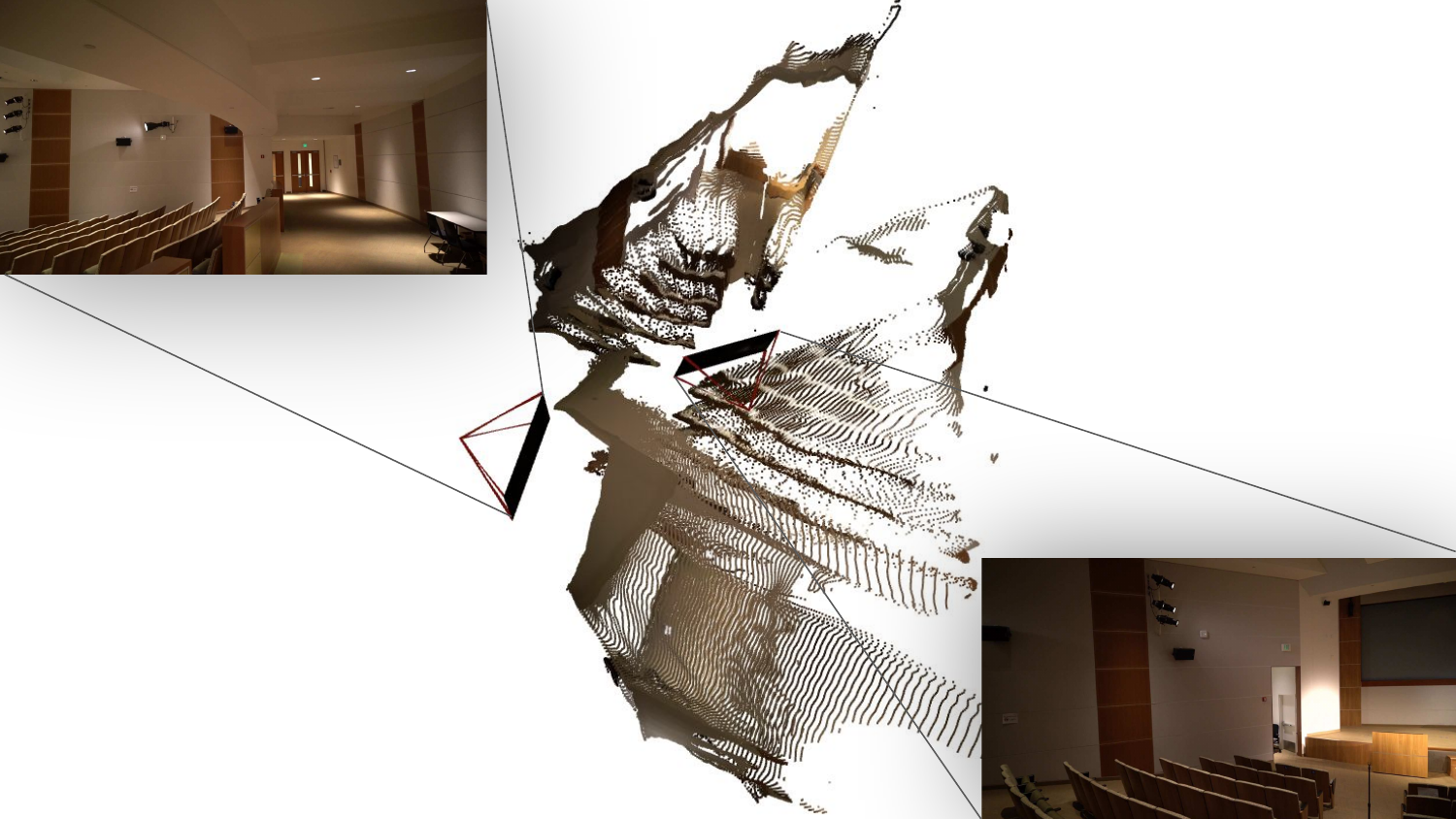}
    \caption{\textbf{Reconstructing opposite-oriented cameras.} After conditioning \method's decoder with the output from our global latent alignment, it is able to predict pointmaps even for images recorded in opposite directions, suggesting the latent global alignment has learned a representation of the entire scene.
    }
    \label{fig:extreme_viewpoints}
    \vspace{-0.5cm}
\end{figure}

%% file: tables/tt_overview_with_mast3r-sfm_values_colored.tex
\begin{tabular}{lllrrrrr}
\toprule
 & Method  & Align. & RRA@5 $\uparrow$ & RTA@5 $\uparrow$ & ATE $\downarrow$ & Reg. $\uparrow$ & Time [s] $\downarrow$ \\
\midrule
\multirow[c]{9}{*}{\verttext{25}} & COLMAP & OPT & {\cellcolor[HTML]{F2C7B9}} \color[HTML]{000000} 13.7 & {\cellcolor[HTML]{EFC1B6}} \color[HTML]{000000} 12.6 & {\cellcolor[HTML]{B7D09D}} \color[HTML]{000000} 0.038 & {\cellcolor[HTML]{E0A4A4}} \color[HTML]{000000} 44.4 & {\cellcolor[HTML]{FFFFFF}} \color[HTML]{000000} - \\
 & GLOMAP & OPT & {\cellcolor[HTML]{CCDAAC}} \color[HTML]{000000} 58.4 & {\cellcolor[HTML]{E6E8BF}} \color[HTML]{000000} 53.6 & {\cellcolor[HTML]{FEE8CC}} \color[HTML]{000000} 0.076 & {\cellcolor[HTML]{E7E8C0}} \color[HTML]{000000} 86.1 & {\cellcolor[HTML]{AACA93}} \color[HTML]{000000} 16.1 \\
 & ACE0 & OPT & {\cellcolor[HTML]{E0A5A4}} \color[HTML]{000000} 1.2 & {\cellcolor[HTML]{E0A4A4}} \color[HTML]{000000} 1.4 & {\cellcolor[HTML]{E0A4A4}} \color[HTML]{000000} 0.112 & {\cellcolor[HTML]{A8C992}} \color[HTML]{000000} 100.0 & {\cellcolor[HTML]{E0A4A4}} \color[HTML]{000000} 1042.7 \\
 & DF-SfM & OPT & {\cellcolor[HTML]{F6F0CB}} \color[HTML]{000000} 47.5 & {\cellcolor[HTML]{F7F0CB}} \color[HTML]{000000} 48.7 & {\cellcolor[HTML]{FFE4CA}} \color[HTML]{000000} 0.081 & {\cellcolor[HTML]{AACA93}} \color[HTML]{000000} 99.4 & {\cellcolor[HTML]{FFFFFF}} \color[HTML]{000000} - \\
 & FlowMap & OPT & {\cellcolor[HTML]{E0A4A4}} \color[HTML]{000000} 0.7 & {\cellcolor[HTML]{E0A4A4}} \color[HTML]{000000} 1.5 & {\cellcolor[HTML]{E5AFAA}} \color[HTML]{000000} 0.107 & {\cellcolor[HTML]{A8C992}} \color[HTML]{000000} 100.0 & {\cellcolor[HTML]{FFFFFF}} \color[HTML]{000000} - \\
 & VGGSfM & OPT & {\cellcolor[HTML]{D6DFB3}} \color[HTML]{000000} 55.7 & {\cellcolor[HTML]{D7E0B4}} \color[HTML]{000000} 57.4 & {\cellcolor[HTML]{F6F0CB}} \color[HTML]{000000} 0.058 & {\cellcolor[HTML]{B8D19E}} \color[HTML]{000000} 96.2 & {\cellcolor[HTML]{C8D8A9}} \color[HTML]{000000} 135.7 \\
 & MASt3R-SfM & OPT & {\cellcolor[HTML]{A8C992}} \color[HTML]{000000} 68.0 & {\cellcolor[HTML]{A8C992}} \color[HTML]{000000} 70.3 & {\cellcolor[HTML]{A8C992}} \color[HTML]{000000} 0.034 & {\cellcolor[HTML]{A8C992}} \color[HTML]{000000} 100.0 & {\cellcolor[HTML]{ECEBC4}} \color[HTML]{000000} 283.2 \\
\cmidrule{3-8}
 & Spann3r & FFD & {\cellcolor[HTML]{FAD7C3}} \color[HTML]{000000} 19.6 & {\cellcolor[HTML]{FEE6CB}} \color[HTML]{000000} 30.7 & {\cellcolor[HTML]{F9F0CD}} \color[HTML]{000000} 0.060 & {\cellcolor[HTML]{A8C992}} \color[HTML]{000000} 100.0 & {\cellcolor[HTML]{A8C992}} \color[HTML]{000000} 8.3 \\
 & \method & FFD & {\cellcolor[HTML]{E9E9C1}} \color[HTML]{000000} 50.9 & {\cellcolor[HTML]{E3E6BD}} \color[HTML]{000000} 54.2 & {\cellcolor[HTML]{D6DFB3}} \color[HTML]{000000} 0.048 & {\cellcolor[HTML]{A8C992}} \color[HTML]{000000} 100.0 & {\cellcolor[HTML]{A8C992}} \color[HTML]{000000} 4.4 \\
\cmidrule{1-8}
\multirow[c]{9}{*}{\verttext{50}} & COLMAP & OPT & {\cellcolor[HTML]{FFE4CA}} \color[HTML]{000000} 28.2 & {\cellcolor[HTML]{FEE2C9}} \color[HTML]{000000} 27.4 & {\cellcolor[HTML]{B7D09D}} \color[HTML]{000000} 0.029 & {\cellcolor[HTML]{E0A4A4}} \color[HTML]{000000} 60.5 & {\cellcolor[HTML]{FFFFFF}} \color[HTML]{000000} - \\
 & GLOMAP & OPT & {\cellcolor[HTML]{A8C992}} \color[HTML]{000000} 69.3 & {\cellcolor[HTML]{A8C992}} \color[HTML]{000000} 70.3 & {\cellcolor[HTML]{EEECC5}} \color[HTML]{000000} 0.039 & {\cellcolor[HTML]{B8D19E}} \color[HTML]{000000} 97.3 & {\cellcolor[HTML]{AECC96}} \color[HTML]{000000} 47.5 \\
 & ACE0 & OPT & {\cellcolor[HTML]{EEBFB4}} \color[HTML]{000000} 11.9 & {\cellcolor[HTML]{EBBAB1}} \color[HTML]{000000} 11.5 & {\cellcolor[HTML]{E3AAA8}} \color[HTML]{000000} 0.071 & {\cellcolor[HTML]{A8C992}} \color[HTML]{000000} 100.0 & {\cellcolor[HTML]{E0A4A4}} \color[HTML]{000000} 1530.0 \\
 & DF-SfM & OPT & {\cellcolor[HTML]{BFD4A3}} \color[HTML]{000000} 63.0 & {\cellcolor[HTML]{C4D6A6}} \color[HTML]{000000} 62.7 & {\cellcolor[HTML]{F7F0CC}} \color[HTML]{000000} 0.041 & {\cellcolor[HTML]{A8C992}} \color[HTML]{000000} 100.0 & {\cellcolor[HTML]{FFFFFF}} \color[HTML]{000000} - \\
 & FlowMap & OPT & {\cellcolor[HTML]{E0A4A4}} \color[HTML]{000000} 1.9 & {\cellcolor[HTML]{E0A4A4}} \color[HTML]{000000} 3.4 & {\cellcolor[HTML]{E0A4A4}} \color[HTML]{000000} 0.073 & {\cellcolor[HTML]{A8C992}} \color[HTML]{000000} 100.0 & {\cellcolor[HTML]{FFFFFF}} \color[HTML]{000000} - \\
 & VGGSfM & OPT & {\cellcolor[HTML]{BED4A2}} \color[HTML]{000000} 63.1 & {\cellcolor[HTML]{BED4A2}} \color[HTML]{000000} 64.2 & {\cellcolor[HTML]{D6DFB3}} \color[HTML]{000000} 0.035 & {\cellcolor[HTML]{B0CD98}} \color[HTML]{000000} 98.7 & {\cellcolor[HTML]{D7E0B4}} \color[HTML]{000000} 291.3 \\
 & MASt3R-SfM & OPT & {\cellcolor[HTML]{A8C992}} \color[HTML]{000000} 69.1 & {\cellcolor[HTML]{A8C992}} \color[HTML]{000000} 70.1 & {\cellcolor[HTML]{A8C992}} \color[HTML]{000000} 0.026 & {\cellcolor[HTML]{A8C992}} \color[HTML]{000000} 100.0 & {\cellcolor[HTML]{F8F0CC}} \color[HTML]{000000} 503.0 \\
\cmidrule{3-8}
 & Spann3r & FFD & {\cellcolor[HTML]{FAD8C4}} \color[HTML]{000000} 21.1 & {\cellcolor[HTML]{FFE6CB}} \color[HTML]{000000} 31.4 & {\cellcolor[HTML]{E5E7BF}} \color[HTML]{000000} 0.037 & {\cellcolor[HTML]{A8C992}} \color[HTML]{000000} 100.0 & {\cellcolor[HTML]{A9C993}} \color[HTML]{000000} 16.7 \\
 & \method & FFD & {\cellcolor[HTML]{E7E8C0}} \color[HTML]{000000} 52.5 & {\cellcolor[HTML]{E1E5BC}} \color[HTML]{000000} 55.2 & {\cellcolor[HTML]{C8D8A9}} \color[HTML]{000000} 0.032 & {\cellcolor[HTML]{A8C992}} \color[HTML]{000000} 100.0 & {\cellcolor[HTML]{A8C992}} \color[HTML]{000000} 8.5 \\
\cmidrule{1-8}
\multirow[c]{9}{*}{\verttext{100}} & COLMAP & OPT & {\cellcolor[HTML]{FCF0CE}} \color[HTML]{000000} 46.4 & {\cellcolor[HTML]{FEEECE}} \color[HTML]{000000} 45.6 & {\cellcolor[HTML]{F9F0CD}} \color[HTML]{000000} 0.026 & {\cellcolor[HTML]{E0A4A4}} \color[HTML]{000000} 85.7 & {\cellcolor[HTML]{FFFFFF}} \color[HTML]{000000} - \\
 & GLOMAP & OPT & {\cellcolor[HTML]{ABCA94}} \color[HTML]{000000} 69.2 & {\cellcolor[HTML]{ADCB96}} \color[HTML]{000000} 71.1 & {\cellcolor[HTML]{EEECC5}} \color[HTML]{000000} 0.024 & {\cellcolor[HTML]{ABCA94}} \color[HTML]{000000} 99.8 & {\cellcolor[HTML]{B3CE9A}} \color[HTML]{000000} 162.3 \\
 & ACE0 & OPT & {\cellcolor[HTML]{FDDDC7}} \color[HTML]{000000} 27.3 & {\cellcolor[HTML]{FDDEC7}} \color[HTML]{000000} 30.6 & {\cellcolor[HTML]{F0C3B7}} \color[HTML]{000000} 0.040 & {\cellcolor[HTML]{A8C992}} \color[HTML]{000000} 100.0 & {\cellcolor[HTML]{E0A4A4}} \color[HTML]{000000} 3393.0 \\
 & DF-SfM & OPT & {\cellcolor[HTML]{B2CE99}} \color[HTML]{000000} 67.4 & {\cellcolor[HTML]{BBD2A0}} \color[HTML]{000000} 67.4 & {\cellcolor[HTML]{FCF0CE}} \color[HTML]{000000} 0.027 & {\cellcolor[HTML]{A9C993}} \color[HTML]{000000} 99.9 & {\cellcolor[HTML]{FFFFFF}} \color[HTML]{000000} - \\
 & FlowMap & OPT & {\cellcolor[HTML]{E0A4A4}} \color[HTML]{000000} 6.8 & {\cellcolor[HTML]{E0A4A4}} \color[HTML]{000000} 10.5 & {\cellcolor[HTML]{E0A4A4}} \color[HTML]{000000} 0.045 & {\cellcolor[HTML]{A8C992}} \color[HTML]{000000} 100.0 & {\cellcolor[HTML]{FFFFFF}} \color[HTML]{000000} - \\
 & VGGSfM & OPT & {\cellcolor[HTML]{CAD9AA}} \color[HTML]{000000} 61.7 & {\cellcolor[HTML]{D3DEB1}} \color[HTML]{000000} 61.8 & {\cellcolor[HTML]{FEEDCE}} \color[HTML]{000000} 0.029 & {\cellcolor[HTML]{C2D5A5}} \color[HTML]{000000} 98.5 & {\cellcolor[HTML]{DAE1B6}} \color[HTML]{000000} 680.0 \\
 & MASt3R-SfM & OPT & {\cellcolor[HTML]{A8C992}} \color[HTML]{000000} 70.1 & {\cellcolor[HTML]{A8C992}} \color[HTML]{000000} 72.3 & {\cellcolor[HTML]{A8C992}} \color[HTML]{000000} 0.017 & {\cellcolor[HTML]{A8C992}} \color[HTML]{000000} 100.0 & {\cellcolor[HTML]{E8E9C1}} \color[HTML]{000000} 861.5 \\
\cmidrule{3-8}
 & Spann3r & FFD & {\cellcolor[HTML]{F8D4C1}} \color[HTML]{000000} 23.5 & {\cellcolor[HTML]{FEE1C9}} \color[HTML]{000000} 32.7 & {\cellcolor[HTML]{F9F0CD}} \color[HTML]{000000} 0.026 & {\cellcolor[HTML]{A8C992}} \color[HTML]{000000} 100.0 & {\cellcolor[HTML]{A9C993}} \color[HTML]{000000} 31.1 \\
 & \method & FFD & {\cellcolor[HTML]{E7E8C0}} \color[HTML]{000000} 54.3 & {\cellcolor[HTML]{EEECC5}} \color[HTML]{000000} 55.2 & {\cellcolor[HTML]{D3DEB1}} \color[HTML]{000000} 0.022 & {\cellcolor[HTML]{A8C992}} \color[HTML]{000000} 100.0 & {\cellcolor[HTML]{A8C992}} \color[HTML]{000000} 16.8 \\
\cmidrule{1-8}
\multirow[c]{9}{*}{\verttext{200}} & COLMAP & OPT & {\cellcolor[HTML]{F7F0CC}} \color[HTML]{000000} 64.7 & {\cellcolor[HTML]{FEECCD}} \color[HTML]{000000} 57.7 & {\cellcolor[HTML]{FDDDC7}} \color[HTML]{000000} 0.019 & {\cellcolor[HTML]{B3CE9A}} \color[HTML]{000000} 97.0 & {\cellcolor[HTML]{FFFFFF}} \color[HTML]{000000} - \\
 & GLOMAP & OPT & {\cellcolor[HTML]{D5DFB2}} \color[HTML]{000000} 73.5 & {\cellcolor[HTML]{D8E0B5}} \color[HTML]{000000} 74.8 & {\cellcolor[HTML]{FEE9CC}} \color[HTML]{000000} 0.016 & {\cellcolor[HTML]{A8C992}} \color[HTML]{000000} 100.0 & {\cellcolor[HTML]{C4D6A6}} \color[HTML]{000000} 536.7 \\
 & ACE0 & OPT & {\cellcolor[HTML]{FEECCD}} \color[HTML]{000000} 55.7 & {\cellcolor[HTML]{FEEBCD}} \color[HTML]{000000} 57.4 & {\cellcolor[HTML]{FDDEC7}} \color[HTML]{000000} 0.019 & {\cellcolor[HTML]{A8C992}} \color[HTML]{000000} 100.0 & {\cellcolor[HTML]{E0A4A4}} \color[HTML]{000000} 4604.4 \\
 & DF-SfM & OPT & {\cellcolor[HTML]{F0EDC6}} \color[HTML]{000000} 66.8 & {\cellcolor[HTML]{EFECC6}} \color[HTML]{000000} 69.3 & {\cellcolor[HTML]{FEE8CC}} \color[HTML]{000000} 0.016 & {\cellcolor[HTML]{E0A4A4}} \color[HTML]{000000} 33.3 & {\cellcolor[HTML]{FFFFFF}} \color[HTML]{000000} - \\
 & FlowMap & OPT & {\cellcolor[HTML]{E0A4A4}} \color[HTML]{000000} 22.2 & {\cellcolor[HTML]{E0A4A4}} \color[HTML]{000000} 25.8 & {\cellcolor[HTML]{E0A4A4}} \color[HTML]{000000} 0.024 & {\cellcolor[HTML]{A8C992}} \color[HTML]{000000} 100.0 & {\cellcolor[HTML]{FFFFFF}} \color[HTML]{000000} - \\
 & VGGSfM & OPT & {\cellcolor[HTML]{A8C992}} \color[HTML]{000000} 84.5 & {\cellcolor[HTML]{A8C992}} \color[HTML]{000000} 86.3 & {\cellcolor[HTML]{A8C992}} \color[HTML]{000000} 0.007 & {\cellcolor[HTML]{F4CBBC}} \color[HTML]{000000} 47.6 & {\cellcolor[HTML]{F7F0CC}} \color[HTML]{000000} 1511.6 \\
 & MASt3R-SfM & OPT & {\cellcolor[HTML]{EAEAC2}} \color[HTML]{000000} 68.2 & {\cellcolor[HTML]{F3EEC9}} \color[HTML]{000000} 68.4 & {\cellcolor[HTML]{F9F0CD}} \color[HTML]{000000} 0.013 & {\cellcolor[HTML]{A8C992}} \color[HTML]{000000} 100.0 & {\cellcolor[HTML]{F9F0CD}} \color[HTML]{000000} 1609.0 \\
\cmidrule{3-8}
 & Spann3r & FFD & {\cellcolor[HTML]{E1A5A5}} \color[HTML]{000000} 22.8 & {\cellcolor[HTML]{E4ACA8}} \color[HTML]{000000} 28.6 & {\cellcolor[HTML]{FBDAC5}} \color[HTML]{000000} 0.019 & {\cellcolor[HTML]{A8C992}} \color[HTML]{000000} 100.0 & {\cellcolor[HTML]{A9C993}} \color[HTML]{000000} 60.4 \\
 & \method & FFD & {\cellcolor[HTML]{FEE9CC}} \color[HTML]{000000} 52.4 & {\cellcolor[HTML]{FEE7CC}} \color[HTML]{000000} 53.1 & {\cellcolor[HTML]{FEEBCD}} \color[HTML]{000000} 0.016 & {\cellcolor[HTML]{A8C992}} \color[HTML]{000000} 100.0 & {\cellcolor[HTML]{A8C992}} \color[HTML]{000000} 33.4 \\
\cmidrule{1-8}
\multirow[c]{9}{*}{\verttext{full}} & COLMAP & OPT & {\cellcolor[HTML]{FFFFFF}} \color[HTML]{000000} GT & {\cellcolor[HTML]{FFFFFF}} \color[HTML]{000000} GT & {\cellcolor[HTML]{FFFFFF}} \color[HTML]{000000} GT & {\cellcolor[HTML]{FFFFFF}} \color[HTML]{000000} GT & {\cellcolor[HTML]{FFFFFF}} \color[HTML]{000000} - \\
 & GLOMAP & OPT & {\cellcolor[HTML]{A8C992}} \color[HTML]{000000} 75.8 & {\cellcolor[HTML]{A8C992}} \color[HTML]{000000} 76.7 & {\cellcolor[HTML]{A8C992}} \color[HTML]{000000} 0.010 & {\cellcolor[HTML]{A8C992}} \color[HTML]{000000} 100.0 & {\cellcolor[HTML]{FAF0CD}} \color[HTML]{000000} 1977.7 \\
 & ACE0 & OPT & {\cellcolor[HTML]{F9F0CD}} \color[HTML]{000000} 56.9 & {\cellcolor[HTML]{FBF0CE}} \color[HTML]{000000} 57.9 & {\cellcolor[HTML]{F9D6C3}} \color[HTML]{000000} 0.015 & {\cellcolor[HTML]{A8C992}} \color[HTML]{000000} 100.0 & {\cellcolor[HTML]{E0A4A4}} \color[HTML]{000000} 5499.5 \\
 & DF-SfM & OPT & {\cellcolor[HTML]{C4D6A6}} \color[HTML]{000000} 69.6 & {\cellcolor[HTML]{CCDAAC}} \color[HTML]{000000} 69.3 & {\cellcolor[HTML]{FEE8CC}} \color[HTML]{000000} 0.014 & {\cellcolor[HTML]{E4E7BE}} \color[HTML]{000000} 76.2 & {\cellcolor[HTML]{FFFFFF}} \color[HTML]{000000} - \\
 & FlowMap & OPT & {\cellcolor[HTML]{F3C9BB}} \color[HTML]{000000} 31.7 & {\cellcolor[HTML]{F4CBBC}} \color[HTML]{000000} 35.7 & {\cellcolor[HTML]{E0A4A4}} \color[HTML]{000000} 0.017 & {\cellcolor[HTML]{F8F0CC}} \color[HTML]{000000} 66.7 & {\cellcolor[HTML]{FFFFFF}} \color[HTML]{000000} - \\
 & VGGSfM & OPT & {\cellcolor[HTML]{FFFFFF}} \color[HTML]{000000} - & {\cellcolor[HTML]{FFFFFF}} \color[HTML]{000000} - & {\cellcolor[HTML]{FFFFFF}} \color[HTML]{000000} - & {\cellcolor[HTML]{E0A4A4}} \color[HTML]{000000} 0.0 & {\cellcolor[HTML]{FDF0CF}} \color[HTML]{000000} 2134.2 \\
 & MASt3R-SfM & OPT & {\cellcolor[HTML]{FEEBCD}} \color[HTML]{000000} 49.2 & {\cellcolor[HTML]{FEEDCE}} \color[HTML]{000000} 54.0 & {\cellcolor[HTML]{BDD3A2}} \color[HTML]{000000} 0.011 & {\cellcolor[HTML]{A8C992}} \color[HTML]{000000} 100.0 & {\cellcolor[HTML]{FEEBCD}} \color[HTML]{000000} 2723.1 \\
\cmidrule{3-8}
 & Spann3r & FFD & {\cellcolor[HTML]{E0A4A4}} \color[HTML]{000000} 20.3 & {\cellcolor[HTML]{E0A4A4}} \color[HTML]{000000} 24.7 & {\cellcolor[HTML]{F1C5B8}} \color[HTML]{000000} 0.016 & {\cellcolor[HTML]{A8C992}} \color[HTML]{000000} 100.0 & {\cellcolor[HTML]{AACA93}} \color[HTML]{000000} 116.2 \\
 & \method & FFD & {\cellcolor[HTML]{FEEECE}} \color[HTML]{000000} 52.0 & {\cellcolor[HTML]{FEECCD}} \color[HTML]{000000} 52.8 & {\cellcolor[HTML]{CFDCAE}} \color[HTML]{000000} 0.011 & {\cellcolor[HTML]{A8C992}} \color[HTML]{000000} 100.0 & {\cellcolor[HTML]{A8C992}} \color[HTML]{000000} 63.4 \\
\bottomrule
\end{tabular}

%% file: tables/ours_vs_spann3r_new.tex
\begin{tabular}{lllrrrr}
\toprule
 & Model & Images  & RRA@5 $\uparrow$ & RTA@5 $\uparrow$ & ATE $\downarrow$ & Time [s] $\downarrow$ \\
\midrule
\multirow[c]{5}{*}{\verttext{25}} & \multirow[c]{3}{*}{Spann3R} & sorted & 19.6 & 30.7 & 0.060 & 8.3 \\
 &  & unordered & 10.6 & 20.1 & 0.070 & 9.3 \\
 &  & all pairs & 20.5 & 31.8 & 0.057 & 77.7 \\
\cmidrule{2-7}
 & \method 224 & SPT & 29.9 & 33.0 & 0.066 & \bfseries 2.2 \\
 & \method & SPT & \bfseries 50.9 & \bfseries 54.2 & \bfseries 0.048 & 4.4 \\
\cmidrule{1-7}
\multirow[c]{5}{*}{\verttext{50}} & \multirow[c]{3}{*}{Spann3R} & sorted & 21.1 & 31.4 & 0.037 & 16.7 \\
 &  & unordered & 12.4 & 19.0 & 0.050 & 18.3 \\ 
 &  & all pairs & 25.8 & 33.5 & 0.043 & 306.0 \\
\cmidrule{2-7}
 & \method 224 & SPT & 34.8 & 36.3 & 0.044 & \bfseries 4.3 \\
 & \method & SPT & \bfseries 52.5 & \bfseries 55.2 & \bfseries 0.032 & 8.5 \\
\cmidrule{1-7}
\multirow[c]{5}{*}{\verttext{full}} & \multirow[c]{3}{*}{Spann3R} & sorted & 20.3 & 24.7 & 0.016 & 116.2 \\
 &  & unordered & 12.8 & 19.8 & 0.018 & 125.6 \\
 &  & all pairs & \multicolumn{4}{c}{OOM} \\
\cmidrule{2-7}
 & \method 224 & SPT & 32.9 & 34.5 & 0.017 & \bfseries 29.4 \\
 & \method & SPT & \bfseries 52.0 & \bfseries 52.8 & \bfseries 0.011 & 63.4 \\
\bottomrule
\end{tabular}

%% file: tables/co3d_10views.tex
\begin{tabular}{llcccc}
\toprule

& \multirow{2}{*}{Method}  & Global& \multicolumn{3}{c}{Co3Dv2$\uparrow$} \\ 
\cline{4-6} 
                      &  & Align. & RRA@15  & RTA@15 & mAA@30 \\ 
\midrule
\multirow{11}{*}{\begin{sideways} 10 \end{sideways}} 
&Colmap~\cite{schonberger2016colmap}        & OPT & 31.6    & 27.3   & 25.3              \\
&Glomap~\cite{pan2024glomap}                & OPT & 45.9    & 40.3   & 37.3     \\
&PixSfM~\cite{lindenberger2021pixsfm}       & OPT & 33.7    & 32.9   & 30.1              \\
&VGGSfM~\cite{wang2024vggsfm}               & OPT & 92.1 & 88.3 & 74.0 \\
&\dustr-GA~\cite{wang2024dust3r}            & OPT & \textbf{96.2}    & 86.8   & 76.7    \\ 
&\mastr-SfM~\cite{duisterhof2024mast3rsfm}  & OPT & 96.0 & \textbf{93.1}   &  \textbf{88.0}   \\
\cline{2-6}
&PoseDiff ~\cite{wang2023posediffusion}     & FFD & 80.5    & 79.8   & 66.5    \\
&PosReg~\cite{wang2023posediffusion}        & FFD & 53.2    & 49.1   & 45.0     \\
&RelPose++~\cite{lin2023relpose++}          & FFD & 82.3    &  77.2      & 65.1   \\
&\spannr~\cite{wang2024spann3r}             & FFD & 89.5 & 83.2 & 70.3    \\
&\mastr\textbf{*}~\cite{leroy2024mast3r}          & FFD &  94.5    &  80.9      & 68.7   \\
&\textbf{\method}                           & FFD & 94.7 & 85.8  &  72.8    \\  %
\midrule
\midrule
\multirow{4}{*}{\begin{sideways} 2 \end{sideways}} 
& \dustr~\cite{wang2024dust3r}              & FFD  & 94.3    & 88.4  & 77.2 \\
& \mastr~\cite{leroy2024mast3r}             & FFD & 94.6   & 91.9   &  \textbf{81.8} \\ 
& \spannr~\cite{wang2024spann3r}            & FFD & 91.9 & 89.9 & 77.6 \\
& \textbf{\method}                          & FFD & \textbf{95.5}  & \textbf{93.2} & 81.6  \\
\bottomrule\\
\end{tabular}

%% file: tables/waymo.tex
\begin{tabular}{lccccc}
\toprule
 &   \multicolumn{4}{c}{Waymo~\cite{waymo_dataset} Val. Split} \\
 \cmidrule{2-5}
Method &   RRA@5$\uparrow$ &  RTA@5$\uparrow$ & ATE$\downarrow$ & Runtime(s)$\downarrow$  \\
\midrule
\mastr-SfM~\cite{leroy2024mast3r} & 75.7 & \textbf{63.7} & \textbf{0.005} & 1662.0 \\
\spannr~\cite{wang2024spann3r}  & 55.1 & 14.5 & 0.025 & 53.8 \\
\textbf{\method}                &  \textbf{78.3} & 57.7 & 0.019 & \textbf{8.5} \\
\bottomrule
\end{tabular}

%% file: tables/ablat_model.tex
\begin{tabular}{clcccccc}
\toprule
 & Backbone & Global & Latent & Graph  &  \multicolumn{3}{c}{Tanks\&Temple~\cite{Knapitsch2017TanksAndTemples} - 200 Views } \\
 \cline{6-8}
 & Init.    & Sup.   & Align. & Const. & RRA@5$\uparrow$ & RTA@5$\uparrow$ & ATE$\downarrow$ \\
\midrule
\multirow{3}{*}{\textbf{(a)}} 
&\mastr  & \xmark & \xmark  & SPT  & 47.5 & 48.3 & 0.019  \\
&\mastr  & \xmark & \cmark  & SPT  & 50.8 &	48.7 & 0.016 \\
&\dustr  & \cmark & \cmark  & SPT  & 48.8 &	48.8 & 0.016  \\
\midrule
\multirow{2}{*}{\textbf{(b)}}
&\mastr  & \cmark & \cmark  & \textit{Oracle} & \textit{52.8} &	\textit{53.8} &	\textit{0.016} \\
&\mastr  & \cmark & \cmark  & MST & 44.4 & 39.5 &0.017 \\	
\midrule
&\mastr  & \cmark & \cmark  & SPT & \textbf{52.4} & \textbf{53.1} &	\textbf{0.016} \\
\bottomrule
\end{tabular}

%% file: tables/pointmap_conf_analysis.tex
\begin{tabular}{lcccccc}
\toprule
 & Conf. thr. & Reg. ($\uparrow$) & RRA@5 ($\uparrow$) & RTA@5 ($\uparrow$) & RRA@15 ($\uparrow$) & RTA@15 ($\uparrow$) \\
\midrule
MASt3R-SfM & N/A & \textbf{100.0} & \textbf{68.0} & \textbf{70.3} & 73.8 & 77.3 \\
\cmidrule{1-2}
\multirow[c]{3}{*}{\textbf{\method}} & 3 & 84.8 & 56.5 & 58.9 & 77.6 & 76.4 \\
 & 5 & 83.2 & 63.0 & 62.2 & 80.0 & 78.7 \\
 & 7 & 75.8 & 65.2 & 63.7 & \textbf{81.3} & \textbf{80.2} \\
\bottomrule
\end{tabular}

%% file: tables/runtime.tex
\begin{tabular}{lcccccrr}
\toprule
Image  & Image &  Latent.   & Graph  & Pointmap & Global & Total & Max. GPU  \\
Resol. & Encoding &  Alignment & Const. & Decoding & Accum. & Runtime(s) & VRAM (GB) \\
 \midrule
224 & 3.6  &  3.4  & 0.1  & 23.2 & 0.9 & 52.3 & 8.0 \\
512 & 12.3  & 7.5  & 1.4  & 68.4 & 1.0 & 135.8 & 25.6  \\

\bottomrule
\end{tabular}

%% file: sec/5_conclusion.tex
\section{Conclusion}
\label{sec:conclusion}

We presented \method, a novel pipeline to perform SfM without traditional components such as matching or global optimization. For this, we build upon 3D foundation models operating on image pairs and scale these to large image collections via a scalable global latent alignment module, effectively aligning pairwise predictions in latent space, replacing global optimization. Further, we leverage a sparse scene graph keeping memory requirements low. We show that such an approach allows to significantly reduce runtime while providing competitive accuracy, opening up exciting new research opportunities towards data-driven approaches for a field that is traditionally dominated by optimization-based methods.

\PAR{Limitations.} We acknowledge that our current model does not scale to all SfM settings, for example for collections of tens of thousands of images. Furthermore, the accuracy of poses at tight thresholds still lacks behind SOTA optimization-based methods, most likely due to the low image resolution processed by learned methods.

%% file: sec/6_supplementary.tex
\newpage
In this supplementary document, we provide implementation details in \cref{sec:supp_impl}, additional evaluations on Tanks\&Temples for pose estimation (\cref{sec:supp_tat}) and Waymo Open Dataset for 3D reconstruction (\cref{sec:supp_3d_recon}).  
Furthermore, we conduct more ablation studies to validate our model design in \cref{sec:supp_ablation}, followed by qualitative visualizations in \cref{sec:supp_visual}.

\section{Implementation Details}
\label{sec:supp_impl}
\PAR{Model.}  Our model adopts the same encoder and decoder architecture as \dustr, \ie, the VIT-L image encoder and the two pointmap decoding regression heads parameterized by VIT-B \cite{dosovitskiyImageWorth16x162020}. For global alignment, we use $L=4$ blocks. $\mathtt{Self}$ and $\mathtt{Cross}$ are implemented as vanilla self-and cross-attention layers~\cite{vaswaniAttentionAllYou2017} with 8 attention heads and pre-normalization. Their feature dimensionality is the same as the VIT-L encoder dimension, \ie, $1024$.

\PAR{Training.}
We train our model on four datasets: Waymo Open Dataset~\cite{waymo_dataset}, CO3Dv2~\cite{reizensteinCommonObjects3D2021}, MegaDepth~\cite{liMegaDepthLearningSingleView2018}, and TartanAir~\cite{wangTartanAirDatasetPush2020}. 
For training, we sample graphs of $N=8$ images based on pairwise scores proposed in CroCo~\cite{weinzaepfelCroCoV2Improved2023} and a greedy algorithm which iteratively adds additional images with maximum viewpoint angle difference \wrt all images already in the set, until the desired number of images is reached. Images are resized such that their longer side has length $512$ and then center cropped such that the shorter side is in $\{384, 336, 288, 256, 160\}$ leading to different aspect ratios for training. Further, we apply color jitter augmentation.
We initialize our model encoder and decoder using \mastr pretrained weights.
We train it for 100,000 iterations with batch size 8 (each batch element corresponds to one graph of images) using AdamW~\cite{loshchilovDecoupledWeightDecay2019} with learning rate $10^{-6}$ and weight decay $5\times 10^{-4}$ on 8 NVIDIA A100-80GB GPUs. The model on small resolutions (using $224\times224)$ is trained on 16 NVIDIA V100-32 GPUs with per-GPU batch size of 2, resulting in overall batch size of 32. We scale the learning rale linearly with batch size.

\PAR{Inference.} At test time, we extract the global camera pose from the pointmaps in global reference frame $X^i$ and their corresponding confidence maps $C^i$. We follow \citet{wang2024dust3r} and first estimate the focal length with a robust estimator~\cite{weiszfeldPointPourLequel1937} and then proceed to extract the pose with RANSAC-PnP~\cite{fischlerRandomSampleConsensus1981,terzakisConsistentlyFastGlobally2020} from points with their corresponding confidence in $C^i$ larger than a threshold. By default we use a threshold of $3$, or the $90$\%-quantile if all confidences fall below the threshold. 

To reduce regression noise, we further symmetrize the edges during inference and combine the pointmap predictions using a confidence-weighted average. In detail, we decode the symmetric edge $(j, i)$, now predicting in the reference frame of image $j$, for every edge $(i, j) \in \espt$,  extract the pairwise pose using Procrustes as described in the main paper, then apply the transformation to the output pointmaps $X^{j, i}$, $X^{j, j}$ to obtain $\tilde{X}^{i, i}$ and $\tilde{X}^{i, j}$ respectively. We then compute the confidence-weighted average for the pointmaps of the edge $(i, j)$ we are interested in.
Here, we introduce the computation to combine $X^{i, i}$ and $\tilde{X}^{i, i}$ but it applies symmetrically to $X^{i, j}$.
First, we compute weight from the confidences $C^{i, i}$ corresponding to edge $(i, j)$ and $C^{j, i}$ from edge $(j, i)$ as 
\begin{equation*}
    G^{i, i}_{u, v} = \frac{\log C^{i, i}_{u, v}}{\log C^{i, i}_{u, v} + \log C^{j, i}_{u, v}}
\end{equation*}
where $u \in \{1, \dots, W\}, v \in \{1, \dots, H\}$ are indexing into the confidence-/pointmaps. Note that the confidence maps correspond to the same image fed in different position to the pairwise decoder.
We then compute the average-weighted pointmap as
\begin{equation*}
    X^{i, i}_{u, v} \coloneqq (G^{i, i}_{u, v}) X^{i, i}_{u, v} \cdot (1 - G^{i, i}_{u, v}) \tilde{X}^{i, i}_{u, v}
\end{equation*}
incorporating information from the decoder evaluation of the symmetric edge, thus refining the pointmap.

\section{Additional Details on Tanks \& Temples~\cite{Knapitsch2017TanksAndTemples}}
\label{sec:supp_tat}

\PAR{Runtime evaluation.}
For completeness, we report the per-scene reconstruction runtime for all baseline methods in \cref{tab:tt_runtimes_all}.
For fair comparison, we run other methods using their open-source implementation with default parameters provided with the code on the same base system with 10 CPU cores, 64GB system memory, and one NVIDIA V100 GPU with 32GB VRAM. 
For \mastr-SfM~\cite{leroy2024mast3r}, we adopt the hyper-parameters reported in the paper. 
We have to do specific adjustment for VGGSfM~\cite{wang2024vggsfm} to fit the GPU memory budget where we follow their suggestions~\footnote{\url{https://github.com/facebookresearch/vggsfm/blob/main/README.md}} 
and reduce \texttt{max\_points\_num} to $40,960$ and \texttt{max\_tri\_points\_num} to $204,800$, \ie, $\nicefrac{1}{4}$ their original values. 
However, this still leads to out-of-memory errors when evaluating on most of the full sequences and some of the 200-image sequences supposedly due to excessive amount of detected keypoints, and thus we do not report the runtime results in these situations.

\begin{figure}[t!]
    \centering
    \includegraphics[width=\linewidth]{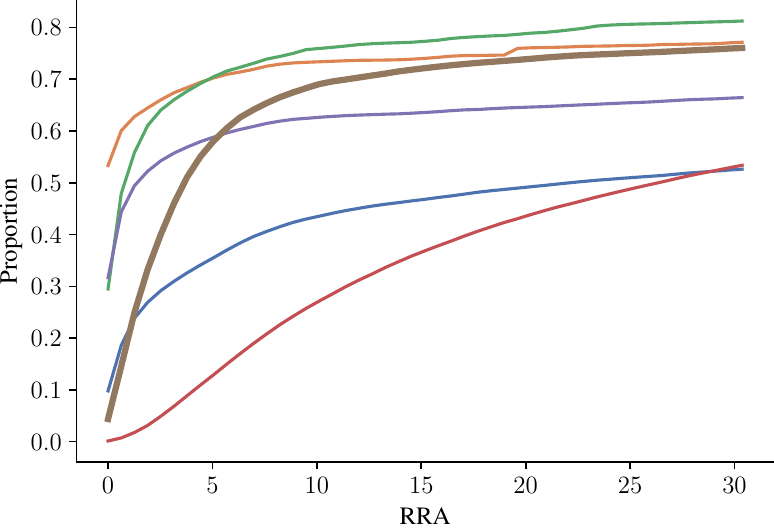}
    \includegraphics[width=\linewidth]{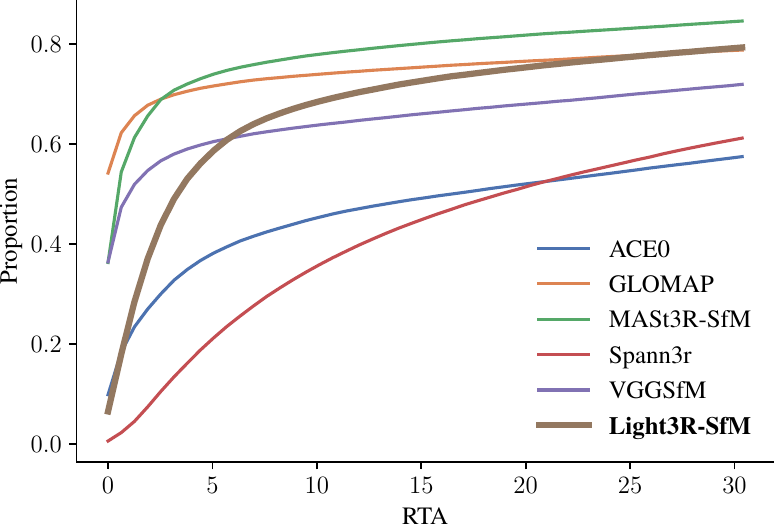}
    \caption{\textbf{CDF of pose errors on 100-view Tanks\&Temples scenes.}}
    \label{fig:cdf_errs}
\end{figure}

\PAR{Pose accuracy evaluation.}
In the main paper, we report pose accuracy at tight error threshold of 5\textdegree. In \cref{fig:cdf_errs}, we provide a more complete overview of the model performance at other thresholds by plotting the pose accuracy as a function of the error threshold for both relative rotation and translation errors.
We observe a gap at tight thresholds between \textit{feed-forward} approaches (\method, \spannr~\cite{wang2024spann3r}) and \textit{optimization-based} approaches, however, this gap rapidly shrinks for our method when moving towards looser thresholds, while \spannr is consistently worse. This suggests that \method is generally able to locate the correct positions and orientations of cameras while struggling to regress the exact values which is more easily achieved via optimization.

For some downstream applications that perform pose refinement, \eg, novel-view synthesis via Gaussian splatting~\cite{kerbl20233dgs}, these coarse poses might already be sufficient and can directly enjoy the significant speed-ups of up to $198\times$ of our method. Further, these results suggest that a small optimization stage on top of the regressed outputs, converging fast due to good initialization, could significantly increase performance at tight thresholds. We leave investigation into this direction to future work.

\section{Evaluation on 3D Reconstruction}
\label{sec:supp_3d_recon}
\begin{figure}[t!]
    \centering
    \includegraphics[width=\linewidth]{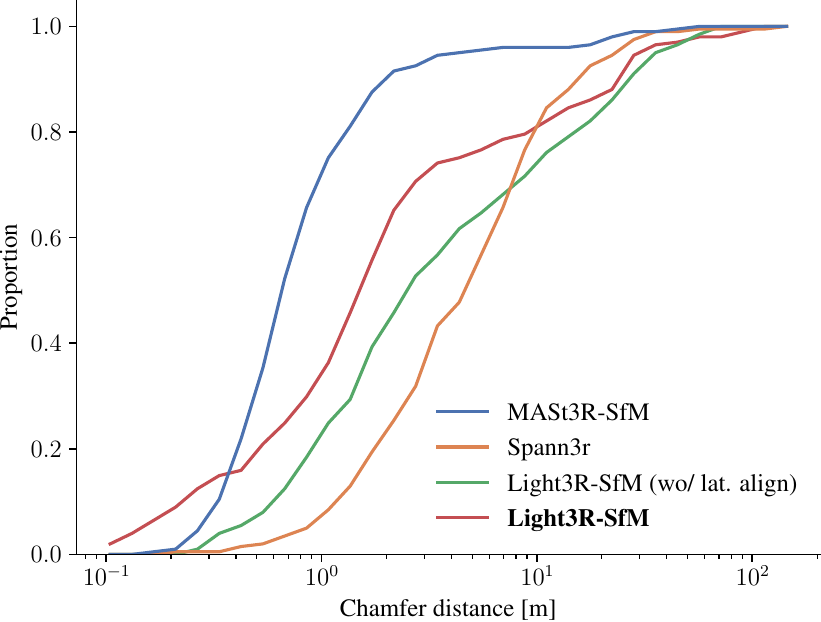}
    \caption{\textbf{CDF of per-scene 3D reconstruction errors.}}
    \label{fig:cdf_3d_reconstruction_errs_waymo}
\end{figure}

We further evaluate our method on 3D reconstruction using Waymo Open Dataset~\cite{waymo_dataset}.
We evaluate the quality of the global predicted point cloud per scene by computing the Chamfer distance~\cite{barrowParametricCorrespondenceChamfer1977} \wrt the sparse lidar ground-truth point cloud. For this, we find the nearest neighbor for every ground truth point and compute the euclidean distance, then compute the average.
We compare ourselves to \spannr and \mastr-SfM, as well as a variant of our method without latent global alignment.
In \cref{fig:cdf_3d_reconstruction_errs_waymo}, we report the cumulative distribution function of per-scene reconstruction errors as measured by the Chamfer distance. 

We show that our methods with and without latent global alignment are both able to largely outperform \spannr, producing point cloud with smaller reconstruction errors for most of the scenes.
It confirms the limitation of \spannr in handling non-object-centric, natural scenes.
We further highlight that our method with latent global alignment module is significantly better than the baseline without it (\textit{wo/ lat.align}), validating its effectiveness to ensure global consistency across pairwise pointmaps, even for the long, forward-moving trajectories.

\begin{figure}[t!]
    \centering
    \includegraphics[width=\linewidth]{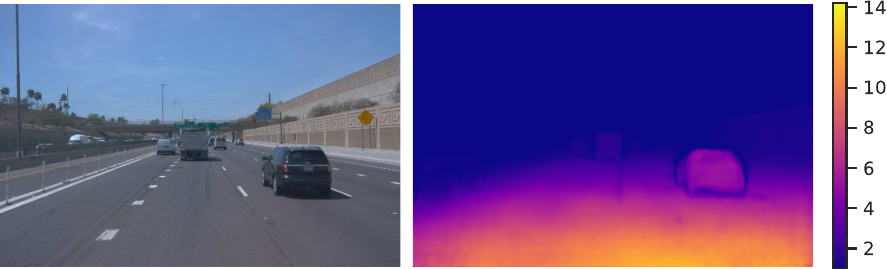}
    \caption{Global confidence map (right) produced by \method for an image of a sequence containing dynamic objects (left).}
    \label{fig:waymo_dynamic_sequence_confidence}
\end{figure}

Compared to the \textit{optimization-based} \mastr-SfM, \method manages to produce a subset of reconstructions with lower reconstruction errors. However, there is also a proportion of scenes where our method falls behind. After investigation, we find that these scenes contain many dynamic objects.
\method was mostly trained on static scenes, and thus often assigns confidence to portions of the pointmap that are dynamic resulting in wrong pairwise pose estimates, affecting global accumulation, and thus degrading global reconstructions. For illustration, we visualize the confidence map for such a dynamic scene in \cref{fig:waymo_dynamic_sequence_confidence}. \mastr-SfM, despite building on top of \mastr as well, performs better in these situations as erroneous correspondences on dynamic objects are discounted by a robust error function during optimization.
We believe \method will be able to handle these scenes by training on more diverse datasets containing dynamic objects, as the global supervision will encourage low confidence for dynamic parts of the image.

\section{Additional Ablation Studies}
\label{sec:supp_ablation}
In addition to the ablation studies performed in the main paper, we consider more detailed ablations for hyper-parameters specific to our contributions. To save compute, we train the models for these ablation studies on lower resolution images, \ie, $224\times224$, versus a maximum resolution of $512\times384$ for the results reported in the main paper. For these experiments, we report results on the 100-view subset of Tanks\&Temples unless otherwise stated.

\PAR{Global alignment layers.}
For results reported in the main paper, we always consider $L=4$ latent global alignment layers. Here we ablate this choice by considering $L \in \{2, 4, 8\}$.
In \cref{tab:ext_abl_num_latent_align_layers}, we report pose accuracy metrics for the different settings of $L$. Using 4 latent alignment layers significantly improves results compared to 2, but doubling the number shows diminishing returns, leading us to select $L=4$ as a trade-off between memory usage/runtime and pose accuracy.

\begin{table}[t!]
    \centering
    \input{tables/supp/abl_num_latent_alignment_layers}
    \caption{\textbf{Impact of number of latent alignment layers $L$.}}
    \label{tab:ext_abl_num_latent_align_layers}
\end{table}

\PAR{Weight of global supervision.}
For the loss supervising the globally aligned pointmaps, accumulated from the pairwise reconstructions, we consider $\lambda=0.1$ as the default. Here, we experiment with other choices of $\lambda$. 

In \cref{tab:ext_abl_global_loss_weight}, we report pose accuracy metrics for choices $\lambda \in \{0.01, 0.1, 1.0\}$. We find that increasing the loss weight from $0.01$ to $0.1$ improves pose estimation, however, the higher setting of $\lambda=1.0$ decreases performance. We explain this behavior with the fact that the global loss produces more noisy supervision compared to the pairwise loss: if a pairwise reconstruction is incorrect it will potentially affects global pointmaps of other views due to global accumulation. Thus, it is beneficial when the pairwise supervision is the main driver of the optimization of model parameters where the global supervision acts as a contributing signal with relative lower weight.

\begin{table}[t!]
    \centering
    \input{tables/supp/abl_global_loss_weight}
    \caption{\textbf{Impact of weight of global supervision $\lambda$.}}
    \label{tab:ext_abl_global_loss_weight}
\end{table}

\begin{table}[t!]
    \centering
    \input{tables/supp/abl_backbone_init}
    \caption{\textbf{Impact of backbone initialization.}}
    \label{tab:ext_abl_initialization}
\end{table}

\PAR{Number of images in training graph.}
All results in the main paper are achieved with models optimized with training graphs of $N=8$ images. In \cref{tab:ext_abl_training_graph_size}, we report results for $N \in \{3, 5, 8, 10\}$. To achieve a fair comparison we increase the batch size for smaller settings of $N$ such that the total number of images per batch and seen over the course of training remains the same. 
Overall, we find a small but consistent improvement for larger training graphs. We explain this consistent improvement by the number of relative constraints in the training graph increasing as the size of the graph increases. With global supervision enforcing consistency of these pairwise constraints the latent alignment layers experience additional supervision leading to better downstream performance. While we achieve better performance on $N=10$, we use $N=8$ for the higher resolution model in the main paper since larger training graphs exceed the GPU memory capacity.

\begin{table}[t!]
    \centering
    \input{tables/supp/abl_training_graph_size}
    \caption{\textbf{Impact of training graph size $N$.} We report results averaged over Tanks\&Temples scenes with all frames.}
    \label{tab:ext_abl_training_graph_size}
\end{table}

\PAR{Model initialization.}
In \cref{tab:ext_abl_initialization}, we report results with different pre-trained weights for the pairwise pointmap regressor used within our method. We find that initializing with either \mastr~\cite{leroy2024mast3r} or the \dustr~\cite{wang2024dust3r} backbone leads to comparable results. If we train the pairwise regressor from scratch, jointly with the other components, we observe that the model performs poorly. This highlights the significance of building on top of geometric foundation models as components for our approach.

\section{Additional Visualizations}
\label{sec:supp_visual}

\begin{figure*}[t!]
    \centering
    \includegraphics[width=\linewidth]{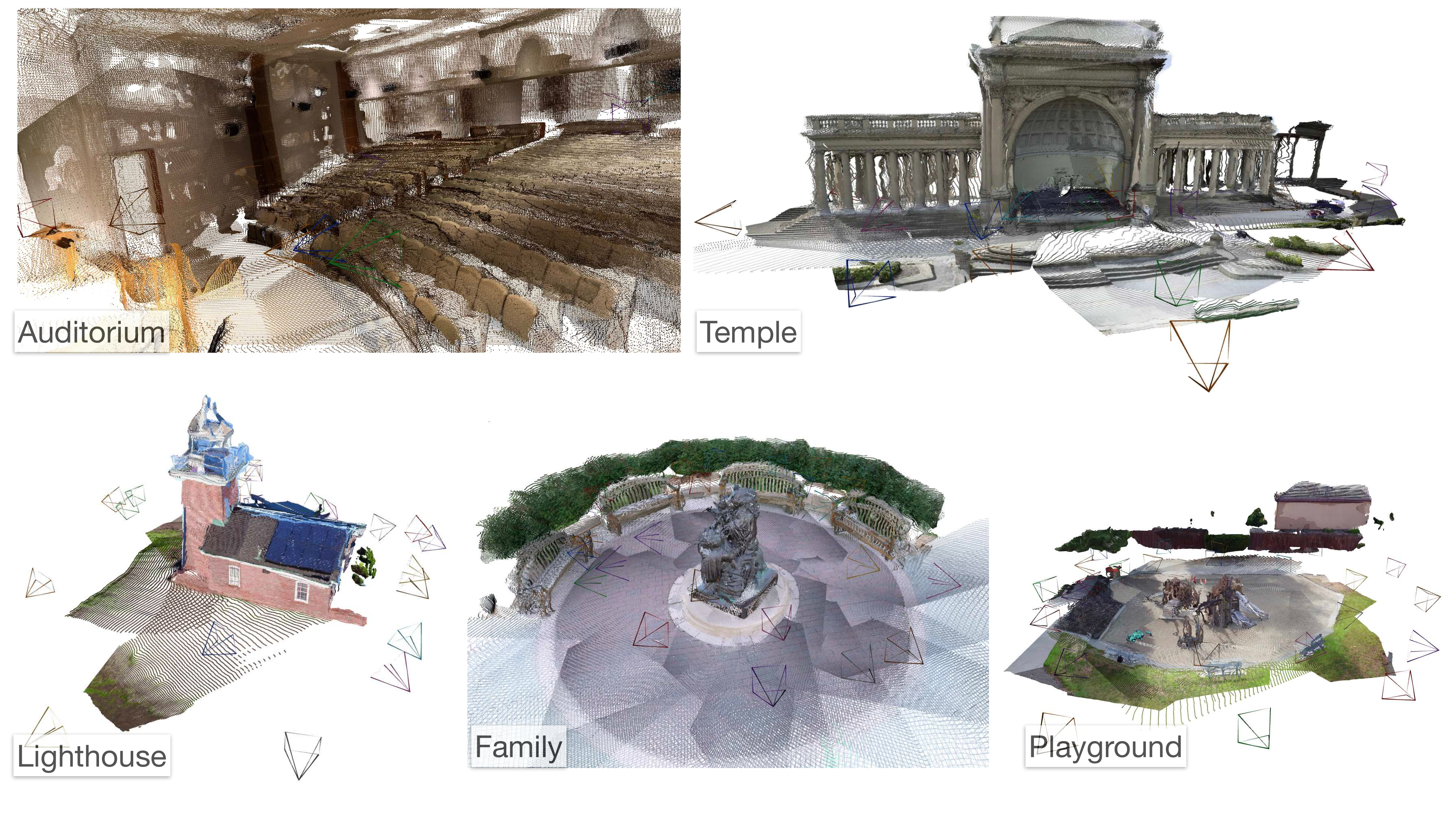}
    \caption{\textbf{Qualitative examples of reconstruction of Tanks \& Temples scenes.}}
    \label{fig:sup:tt_vis}
\end{figure*}

\begin{figure*}[t!]
    \centering
    \includegraphics[width=\linewidth]{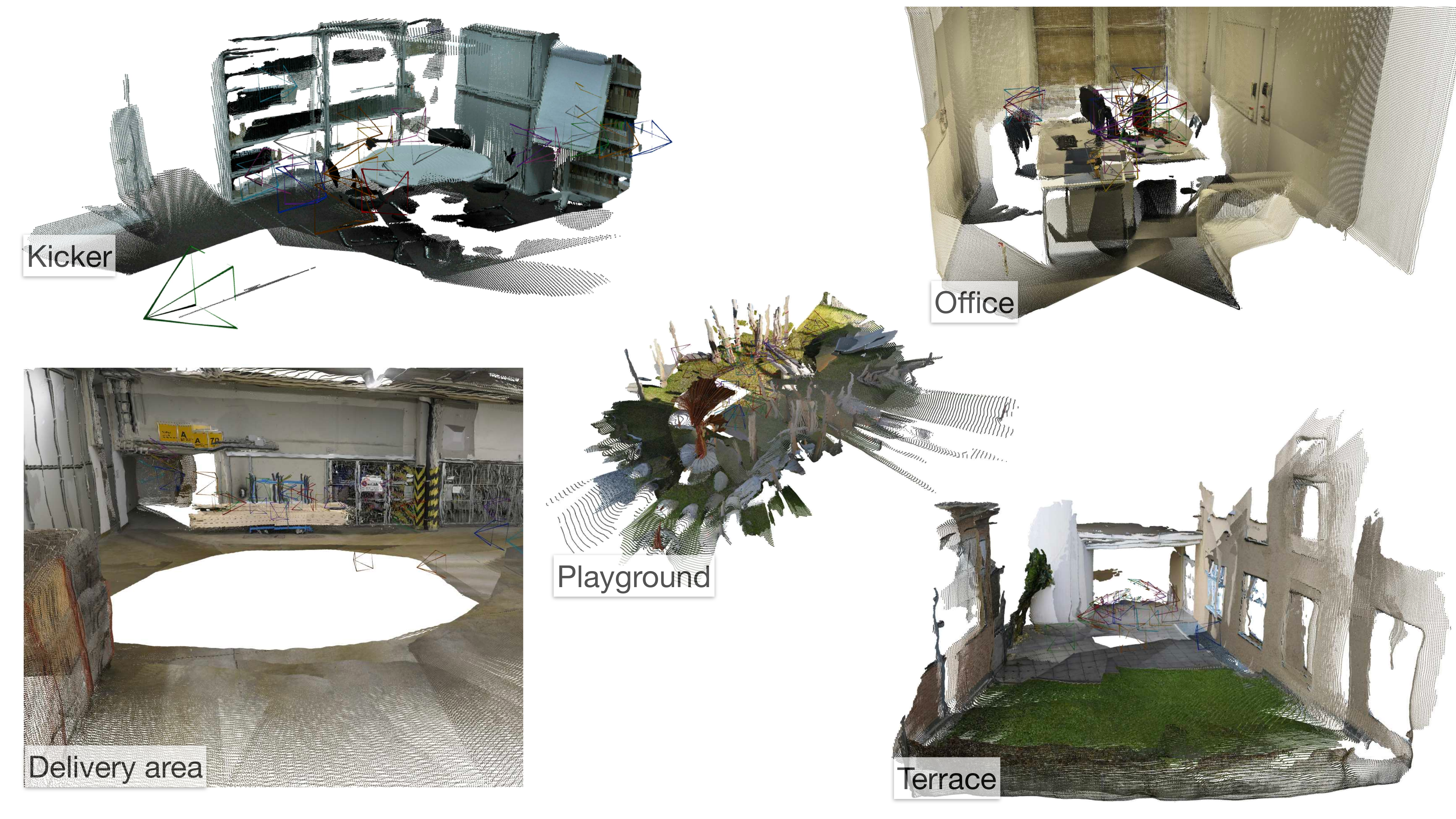}
    \caption{\textbf{Qualitative examples of reconstruction of ETH3D scenes.}}
    \label{fig:sup:eth3d_vis}
\end{figure*}

\PAR{Reconstruction examples.} We provide visualizations of reconstructions obtained using \method. In \cref{fig:sup:tt_vis}, we show reconstruction of diverse Tanks\&Temples scenes, including indoor, object-centric, and large scale reconstructions of landmarks. Further, we provide qualitative results on the challenging ETH3D~\cite{schopsMultiViewStereoBenchmark2017} scenes in \cref{fig:sup:eth3d_vis}. 

\begin{figure*}[t!]
    \centering
    \includegraphics[width=0.8\linewidth]{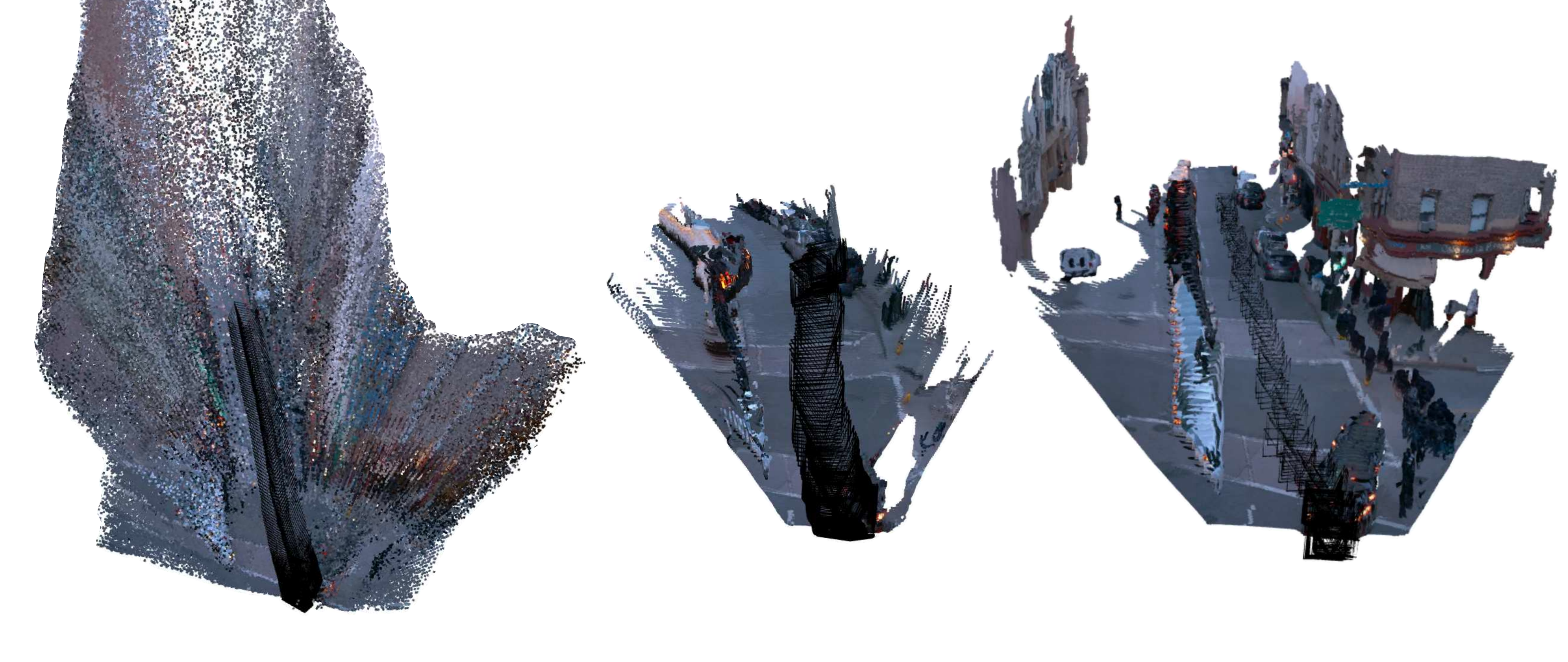}
    \includegraphics[width=0.8\linewidth]{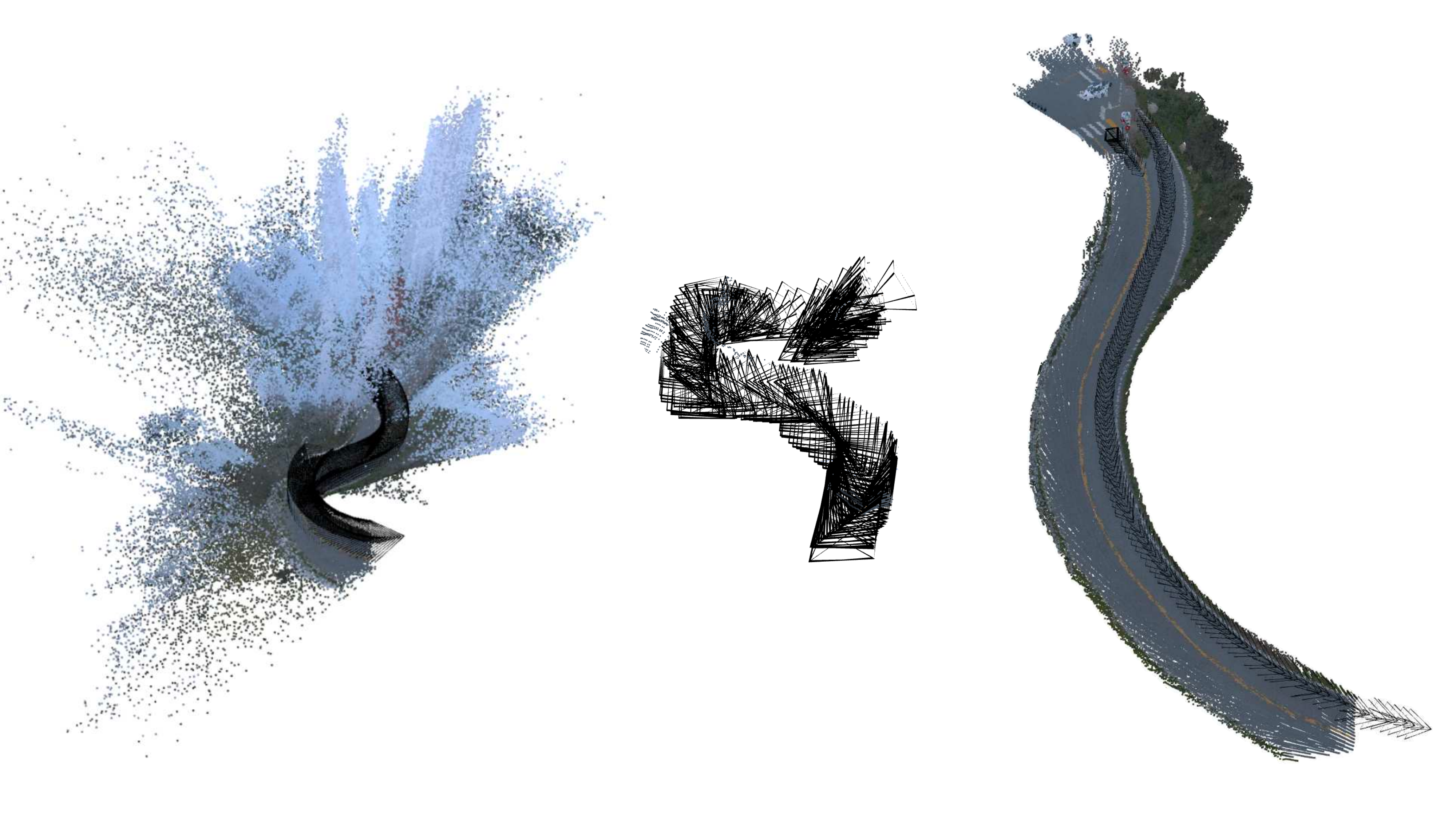}
    \includegraphics[width=0.8\linewidth]{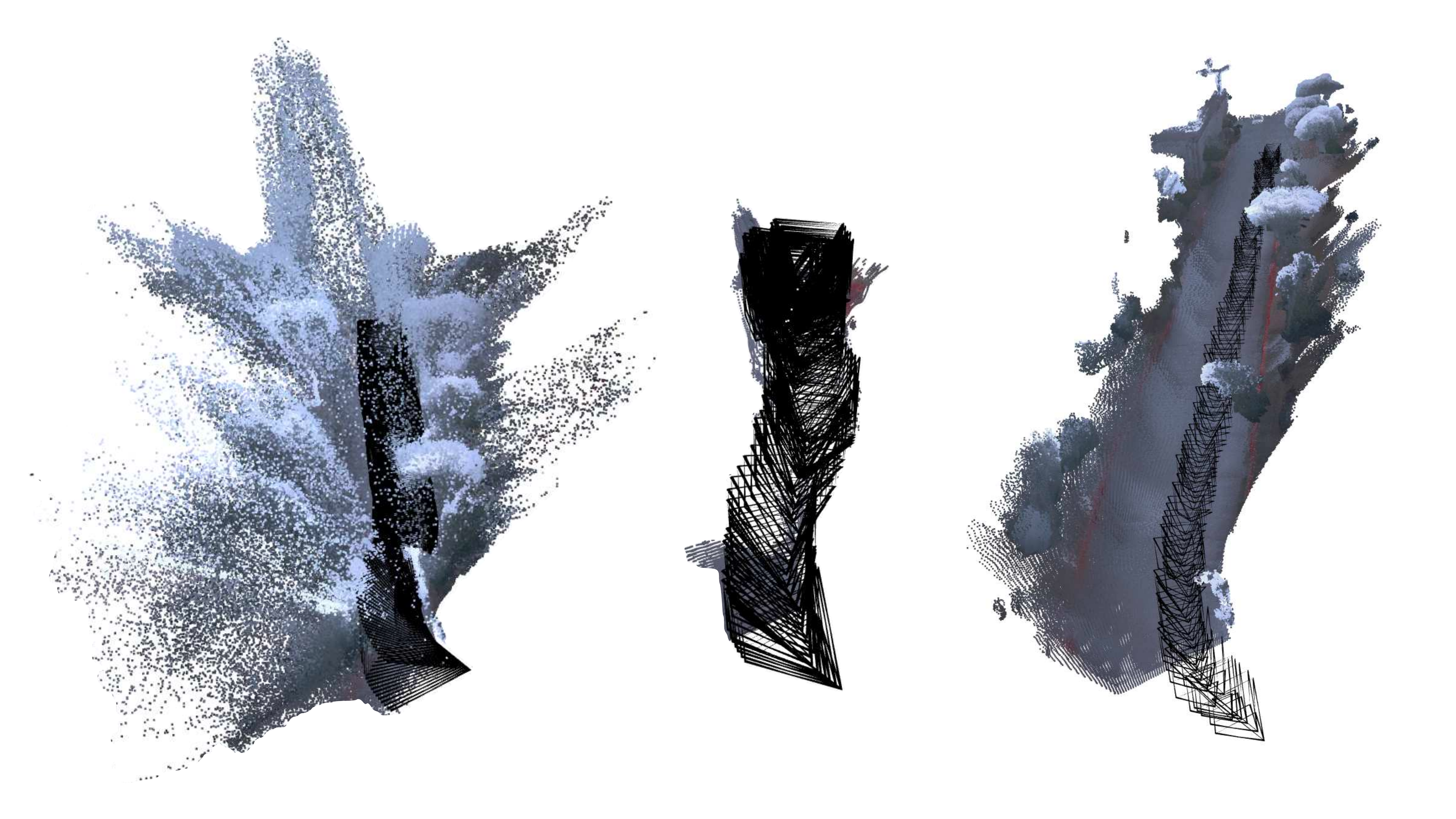}
    \caption{\textbf{More comparisons on Waymo.} Comparing from left-to-right: \mastr-SfM, \spannr, \method.}
    \label{fig:waymo_additional_comp}
\end{figure*}

\PAR{Qualitative comparisons on Waymo sequences.} We provide additional qualitative comparison of 3D reconstructions from the Waymo Open Dataset~\cite{waymo_dataset} obtained by \mastr-SfM, \spannr, and \method.
As shown in \cref{fig:waymo_additional_comp}, \spannr fails to reconstruct the camera poses as well as the scene structure when the trajectory is longer, while \mastr-SfM fails to recover the boundary and further away background regions, leading to noisy and coarse reconstruction. In contrast, our method is able to recover accurate camera poses as well as capture fine details in the scene, \eg, cars and buildings along the street.

\begin{figure*}[t!]
    \centering
    \includegraphics[width=\linewidth]{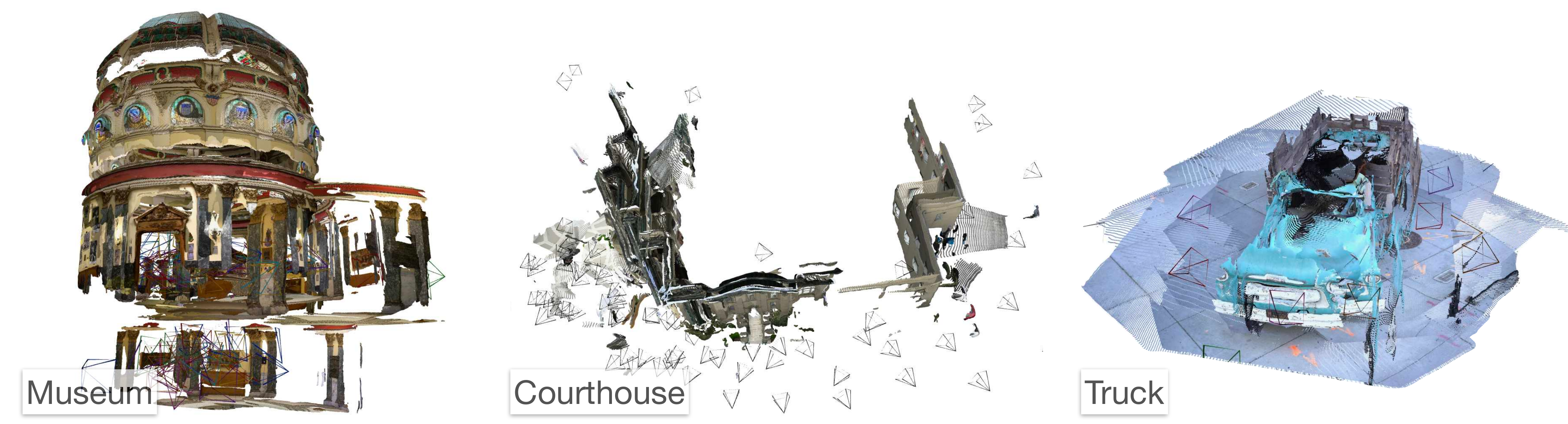}
    \caption{\textbf{Failure cases on the Tanks \& Temples dataset.}}
    \label{fig:sup:tt_failures_vis}
\end{figure*}

\PAR{Failure cases.} Finally, we provide visualizations of typical failure cases in \cref{fig:sup:tt_failures_vis}. We observed that retrieval failures can result in multiple sub-reconstructions which are aligned within themselves but globally inconsistent. Further, small errors in the pairwise estimations result in misalignment in the global reconstruction.

\begin{table*}[t]
    \begin{minipage}[t]{0.48\textwidth} %
        \centering
        \resizebox{\linewidth}{!}{\input{tables/supp/tt_runtimes_all}}
    \end{minipage}%
    \hfill
    \begin{minipage}[t]{0.48\textwidth} %
        \centering
        \resizebox{\linewidth}{!}{\input{tables/supp/tt_runtimes_all_part2}}
    \end{minipage}
    \caption{\textbf{Per-scene reconstruction runtimes on Tanks\&Temples.} All runtimes are reported in seconds.}
    \label{tab:tt_runtimes_all}
\end{table*}

%% file: tables/supp/abl_num_latent_alignment_layers.tex
\begin{tabular}{lccc}
\toprule
 & RRA@5 $\uparrow$ & RTA@5 $\uparrow$ & ATE $\downarrow$ \\
$L$ &  &  &  \\
\midrule
2 & 33.1 & 35.5 & 0.033 \\
4 & \bfseries 35.7 & \bfseries 36.9 & \bfseries 0.032 \\
8 & 35.3 & \bfseries 36.9 & \bfseries 0.032 \\
\bottomrule
\end{tabular}

%% file: tables/supp/abl_global_loss_weight.tex
\begin{tabular}{lccc}
\toprule
 & RRA@5 $\uparrow$ & RTA@5 $\uparrow$ & ATE $\downarrow$ \\
$\lambda$ &  &  &  \\
\midrule
0.01 & 30.6 & 33.3 & 0.034 \\
0.1 & \bfseries 35.7 & \bfseries 36.9 & 0.032 \\
1 & 34.5 & 36.8 & \bfseries 0.031 \\
\bottomrule
\end{tabular}

%% file: tables/supp/abl_backbone_init.tex
\begin{tabular}{lccc}
\toprule
 & RRA@5 $\uparrow$ & RTA@5 $\uparrow$ & ATE $\downarrow$ \\
Backbone init. &  &  &  \\
\midrule
Scratch & 0.7 & 0.1 & 0.057 \\
\dustr & \bfseries 35.7 & 36.9 & \bfseries 0.032 \\
\mastr & 34.6 & \bfseries 38.6 & \bfseries 0.032 \\
\bottomrule
\end{tabular}

%% file: tables/supp/abl_training_graph_size.tex
\begin{tabular}{lccc}
\toprule
 & RRA@5 $\uparrow$ & RTA@5 $\uparrow$ & ATE $\downarrow$ \\
N &  &  &  \\
\midrule
3 & 37.3 & 38.1 & 0.033 \\
5 & 38.0 & 39.4 & \bfseries 0.030 \\
8 & 39.0 & 39.5 & \bfseries 0.030 \\
10 & \bfseries 39.2 & \bfseries 40.7 & 0.031 \\
\bottomrule
\end{tabular}

%% file: tables/supp/tt_runtimes_all.tex
\begin{tabular}{lllcccccc}
\toprule
\verttext{\#images} & \verttext{Group}  & Scene & \verttext{ACE0{~\cite{brachmann2024acezero}}} & \verttext{  MASt3R-SfM~\cite{duisterhof2024mast3rsfm}  } & \verttext{VGGSfM~\cite{wang2024vggsfm}} & \verttext{GLOMAP~\cite{pan2024glomap}} & \verttext{Spann3r~\cite{wang2024spann3r}} & \verttext{\textbf{\method}} \\
\midrule
\multirow[c]{21}{*}{\verttext{25}} & \multirow[c]{6}{*}{\verttext{Advanced}} & Auditorium & {\cellcolor[HTML]{E0A4A4}} \color[HTML]{000000} 1113.5 & {\cellcolor[HTML]{E9E9C1}} \color[HTML]{000000} 288.1 & {\cellcolor[HTML]{BAD2A0}} \color[HTML]{000000} 90.7 & {\cellcolor[HTML]{A9C993}} \color[HTML]{000000} 10.4 & {\cellcolor[HTML]{A9C993}} \color[HTML]{000000} 10.3 & {\cellcolor[HTML]{A8C992}} \color[HTML]{000000} 5.8 \\
 &  & Ballroom & {\cellcolor[HTML]{E0A4A4}} \color[HTML]{000000} 1141.6 & {\cellcolor[HTML]{E9E9C1}} \color[HTML]{000000} 294.3 & {\cellcolor[HTML]{CCDAAC}} \color[HTML]{000000} 167.3 & {\cellcolor[HTML]{ACCB95}} \color[HTML]{000000} 25.4 & {\cellcolor[HTML]{A8C992}} \color[HTML]{000000} 7.6 & {\cellcolor[HTML]{A8C992}} \color[HTML]{000000} 4.3 \\
 &  & Courtroom & {\cellcolor[HTML]{E0A4A4}} \color[HTML]{000000} 1534.4 & {\cellcolor[HTML]{D7E0B4}} \color[HTML]{000000} 290.3 & {\cellcolor[HTML]{BBD2A0}} \color[HTML]{000000} 125.2 & {\cellcolor[HTML]{A9C993}} \color[HTML]{000000} 10.9 & {\cellcolor[HTML]{A8C992}} \color[HTML]{000000} 7.1 & {\cellcolor[HTML]{A8C992}} \color[HTML]{000000} 4.3 \\
 &  & Museum & {\cellcolor[HTML]{E0A4A4}} \color[HTML]{000000} 917.4 & {\cellcolor[HTML]{F6F0CB}} \color[HTML]{000000} 287.0 & {\cellcolor[HTML]{D2DDB0}} \color[HTML]{000000} 155.2 & {\cellcolor[HTML]{ABCA94}} \color[HTML]{000000} 15.5 & {\cellcolor[HTML]{A9C993}} \color[HTML]{000000} 8.7 & {\cellcolor[HTML]{A8C992}} \color[HTML]{000000} 4.4 \\
 &  & Palace & {\cellcolor[HTML]{FFFFFF}} \color[HTML]{000000} - & {\cellcolor[HTML]{E0A4A4}} \color[HTML]{000000} 286.3 & {\cellcolor[HTML]{F6CFBE}} \color[HTML]{000000} 219.7 & {\cellcolor[HTML]{ABCA94}} \color[HTML]{000000} 7.3 & {\cellcolor[HTML]{ABCA94}} \color[HTML]{000000} 7.3 & {\cellcolor[HTML]{A8C992}} \color[HTML]{000000} 3.9 \\
 &  & Temple & {\cellcolor[HTML]{FFFFFF}} \color[HTML]{000000} - & {\cellcolor[HTML]{E0A4A4}} \color[HTML]{000000} 285.0 & {\cellcolor[HTML]{EEC0B5}} \color[HTML]{000000} 241.2 & {\cellcolor[HTML]{ACCB95}} \color[HTML]{000000} 9.4 & {\cellcolor[HTML]{ABCA94}} \color[HTML]{000000} 7.7 & {\cellcolor[HTML]{A8C992}} \color[HTML]{000000} 4.1 \\
\cmidrule{2-9}
 & \multirow[c]{8}{*}{\verttext{Intermediate}} & Family & {\cellcolor[HTML]{E0A4A4}} \color[HTML]{000000} 1334.7 & {\cellcolor[HTML]{DDE3B9}} \color[HTML]{000000} 285.4 & {\cellcolor[HTML]{B7D09D}} \color[HTML]{000000} 83.5 & {\cellcolor[HTML]{AACA93}} \color[HTML]{000000} 20.1 & {\cellcolor[HTML]{A9C993}} \color[HTML]{000000} 10.9 & {\cellcolor[HTML]{A8C992}} \color[HTML]{000000} 5.6 \\
 &  & Francis & {\cellcolor[HTML]{E0A4A4}} \color[HTML]{000000} 970.6 & {\cellcolor[HTML]{F0EDC6}} \color[HTML]{000000} 277.8 & {\cellcolor[HTML]{BBD2A0}} \color[HTML]{000000} 80.1 & {\cellcolor[HTML]{AACA93}} \color[HTML]{000000} 13.0 & {\cellcolor[HTML]{A8C992}} \color[HTML]{000000} 7.5 & {\cellcolor[HTML]{A8C992}} \color[HTML]{000000} 3.9 \\
 &  & Horse & {\cellcolor[HTML]{E0A4A4}} \color[HTML]{000000} 980.2 & {\cellcolor[HTML]{F1EDC7}} \color[HTML]{000000} 287.1 & {\cellcolor[HTML]{B7D09D}} \color[HTML]{000000} 66.9 & {\cellcolor[HTML]{AECC96}} \color[HTML]{000000} 29.3 & {\cellcolor[HTML]{A9C993}} \color[HTML]{000000} 11.8 & {\cellcolor[HTML]{A8C992}} \color[HTML]{000000} 5.5 \\
 &  & Lighthouse & {\cellcolor[HTML]{E0A4A4}} \color[HTML]{000000} 881.5 & {\cellcolor[HTML]{F7F0CC}} \color[HTML]{000000} 282.0 & {\cellcolor[HTML]{C7D8A8}} \color[HTML]{000000} 110.7 & {\cellcolor[HTML]{A9C993}} \color[HTML]{000000} 7.6 & {\cellcolor[HTML]{A9C993}} \color[HTML]{000000} 7.3 & {\cellcolor[HTML]{A8C992}} \color[HTML]{000000} 3.7 \\
 &  & M60 & {\cellcolor[HTML]{E0A4A4}} \color[HTML]{000000} 826.8 & {\cellcolor[HTML]{F7F0CC}} \color[HTML]{000000} 268.0 & {\cellcolor[HTML]{E2E6BC}} \color[HTML]{000000} 191.7 & {\cellcolor[HTML]{ADCB96}} \color[HTML]{000000} 22.3 & {\cellcolor[HTML]{A8C992}} \color[HTML]{000000} 7.0 & {\cellcolor[HTML]{A8C992}} \color[HTML]{000000} 3.8 \\
 &  & Panther & {\cellcolor[HTML]{E0A4A4}} \color[HTML]{000000} 831.1 & {\cellcolor[HTML]{F7F0CC}} \color[HTML]{000000} 268.5 & {\cellcolor[HTML]{D2DDB0}} \color[HTML]{000000} 140.5 & {\cellcolor[HTML]{ABCA94}} \color[HTML]{000000} 14.4 & {\cellcolor[HTML]{A8C992}} \color[HTML]{000000} 7.0 & {\cellcolor[HTML]{A8C992}} \color[HTML]{000000} 4.0 \\
 &  & Playground & {\cellcolor[HTML]{E0A4A4}} \color[HTML]{000000} 866.5 & {\cellcolor[HTML]{F8F0CC}} \color[HTML]{000000} 291.1 & {\cellcolor[HTML]{C3D6A5}} \color[HTML]{000000} 97.6 & {\cellcolor[HTML]{A9C993}} \color[HTML]{000000} 8.0 & {\cellcolor[HTML]{A9C993}} \color[HTML]{000000} 8.3 & {\cellcolor[HTML]{A8C992}} \color[HTML]{000000} 4.4 \\
 &  & Train & {\cellcolor[HTML]{FFFFFF}} \color[HTML]{000000} - & {\cellcolor[HTML]{E0A4A4}} \color[HTML]{000000} 290.5 & {\cellcolor[HTML]{FDF0CF}} \color[HTML]{000000} 114.2 & {\cellcolor[HTML]{B5CF9B}} \color[HTML]{000000} 19.8 & {\cellcolor[HTML]{ABCA94}} \color[HTML]{000000} 7.8 & {\cellcolor[HTML]{A8C992}} \color[HTML]{000000} 4.4 \\
\cmidrule{2-9}
 & \multirow[c]{7}{*}{\verttext{Train}} & Barn & {\cellcolor[HTML]{E0A4A4}} \color[HTML]{000000} 986.6 & {\cellcolor[HTML]{EEECC5}} \color[HTML]{000000} 273.6 & {\cellcolor[HTML]{CBDAAB}} \color[HTML]{000000} 140.9 & {\cellcolor[HTML]{A9C993}} \color[HTML]{000000} 10.4 & {\cellcolor[HTML]{A9C993}} \color[HTML]{000000} 8.8 & {\cellcolor[HTML]{A8C992}} \color[HTML]{000000} 4.4 \\
 &  & Caterpillar & {\cellcolor[HTML]{E0A4A4}} \color[HTML]{000000} 1229.5 & {\cellcolor[HTML]{E2E6BC}} \color[HTML]{000000} 285.4 & {\cellcolor[HTML]{B7D09D}} \color[HTML]{000000} 79.3 & {\cellcolor[HTML]{AACA93}} \color[HTML]{000000} 15.1 & {\cellcolor[HTML]{A8C992}} \color[HTML]{000000} 7.2 & {\cellcolor[HTML]{A8C992}} \color[HTML]{000000} 4.4 \\
 &  & Church & {\cellcolor[HTML]{E0A4A4}} \color[HTML]{000000} 1088.3 & {\cellcolor[HTML]{E6E8BF}} \color[HTML]{000000} 268.4 & {\cellcolor[HTML]{C0D5A4}} \color[HTML]{000000} 111.4 & {\cellcolor[HTML]{ACCB95}} \color[HTML]{000000} 24.0 & {\cellcolor[HTML]{A9C993}} \color[HTML]{000000} 9.9 & {\cellcolor[HTML]{A8C992}} \color[HTML]{000000} 4.4 \\
 &  & Courthouse & {\cellcolor[HTML]{FFFFFF}} \color[HTML]{000000} - & {\cellcolor[HTML]{EAB7AF}} \color[HTML]{000000} 293.0 & {\cellcolor[HTML]{E0A4A4}} \color[HTML]{000000} 328.4 & {\cellcolor[HTML]{B3CE9A}} \color[HTML]{000000} 18.0 & {\cellcolor[HTML]{ABCA94}} \color[HTML]{000000} 8.8 & {\cellcolor[HTML]{A8C992}} \color[HTML]{000000} 3.7 \\
 &  & Ignatius & {\cellcolor[HTML]{E0A4A4}} \color[HTML]{000000} 931.6 & {\cellcolor[HTML]{EFECC6}} \color[HTML]{000000} 262.5 & {\cellcolor[HTML]{BAD2A0}} \color[HTML]{000000} 75.7 & {\cellcolor[HTML]{ABCA94}} \color[HTML]{000000} 16.9 & {\cellcolor[HTML]{A8C992}} \color[HTML]{000000} 6.9 & {\cellcolor[HTML]{A8C992}} \color[HTML]{000000} 4.3 \\
 &  & Meetingroom & {\cellcolor[HTML]{E0A4A4}} \color[HTML]{000000} 1165.0 & {\cellcolor[HTML]{E5E7BF}} \color[HTML]{000000} 280.7 & {\cellcolor[HTML]{C3D6A5}} \color[HTML]{000000} 127.7 & {\cellcolor[HTML]{A9C993}} \color[HTML]{000000} 12.2 & {\cellcolor[HTML]{A9C993}} \color[HTML]{000000} 9.1 & {\cellcolor[HTML]{A8C992}} \color[HTML]{000000} 4.1 \\
 &  & Truck & {\cellcolor[HTML]{E0A4A4}} \color[HTML]{000000} 926.3 & {\cellcolor[HTML]{F7F0CC}} \color[HTML]{000000} 301.8 & {\cellcolor[HTML]{C3D6A5}} \color[HTML]{000000} 102.3 & {\cellcolor[HTML]{AECC96}} \color[HTML]{000000} 27.8 & {\cellcolor[HTML]{A8C992}} \color[HTML]{000000} 8.0 & {\cellcolor[HTML]{A8C992}} \color[HTML]{000000} 4.6 \\
\cmidrule{1-9} \cmidrule{2-9}
\multirow[c]{21}{*}{\verttext{50}} & \multirow[c]{6}{*}{\verttext{Advanced}} & Auditorium & {\cellcolor[HTML]{E0A4A4}} \color[HTML]{000000} 1497.2 & {\cellcolor[HTML]{FAF0CD}} \color[HTML]{000000} 532.4 & {\cellcolor[HTML]{D6DFB3}} \color[HTML]{000000} 282.0 & {\cellcolor[HTML]{ABCA94}} \color[HTML]{000000} 32.5 & {\cellcolor[HTML]{A9C993}} \color[HTML]{000000} 20.5 & {\cellcolor[HTML]{A8C992}} \color[HTML]{000000} 10.5 \\
 &  & Ballroom & {\cellcolor[HTML]{E0A4A4}} \color[HTML]{000000} 1848.6 & {\cellcolor[HTML]{F0EDC6}} \color[HTML]{000000} 529.1 & {\cellcolor[HTML]{CCDAAC}} \color[HTML]{000000} 272.5 & {\cellcolor[HTML]{B6CF9C}} \color[HTML]{000000} 110.9 & {\cellcolor[HTML]{A8C992}} \color[HTML]{000000} 14.7 & {\cellcolor[HTML]{A8C992}} \color[HTML]{000000} 9.0 \\
 &  & Courtroom & {\cellcolor[HTML]{E0A4A4}} \color[HTML]{000000} 1480.7 & {\cellcolor[HTML]{F9F0CD}} \color[HTML]{000000} 505.6 & {\cellcolor[HTML]{EEECC5}} \color[HTML]{000000} 415.5 & {\cellcolor[HTML]{ACCB95}} \color[HTML]{000000} 32.6 & {\cellcolor[HTML]{A9C993}} \color[HTML]{000000} 16.3 & {\cellcolor[HTML]{A8C992}} \color[HTML]{000000} 8.8 \\
 &  & Museum & {\cellcolor[HTML]{E0A4A4}} \color[HTML]{000000} 992.5 & {\cellcolor[HTML]{FEECCD}} \color[HTML]{000000} 468.8 & {\cellcolor[HTML]{FDDEC7}} \color[HTML]{000000} 670.5 & {\cellcolor[HTML]{B2CE99}} \color[HTML]{000000} 48.4 & {\cellcolor[HTML]{AACA93}} \color[HTML]{000000} 17.8 & {\cellcolor[HTML]{A8C992}} \color[HTML]{000000} 8.1 \\
 &  & Palace & {\cellcolor[HTML]{FFFFFF}} \color[HTML]{000000} - & {\cellcolor[HTML]{F2C7B9}} \color[HTML]{000000} 486.1 & {\cellcolor[HTML]{E0A4A4}} \color[HTML]{000000} 601.1 & {\cellcolor[HTML]{AECC96}} \color[HTML]{000000} 21.7 & {\cellcolor[HTML]{ABCA94}} \color[HTML]{000000} 16.3 & {\cellcolor[HTML]{A8C992}} \color[HTML]{000000} 7.6 \\
 &  & Temple & {\cellcolor[HTML]{FFFFFF}} \color[HTML]{000000} - & {\cellcolor[HTML]{E3AAA8}} \color[HTML]{000000} 481.7 & {\cellcolor[HTML]{E0A4A4}} \color[HTML]{000000} 499.9 & {\cellcolor[HTML]{B0CD98}} \color[HTML]{000000} 25.0 & {\cellcolor[HTML]{ACCB95}} \color[HTML]{000000} 16.3 & {\cellcolor[HTML]{A8C992}} \color[HTML]{000000} 8.5 \\
\cmidrule{2-9}
 & \multirow[c]{8}{*}{\verttext{Intermediate}} & Family & {\cellcolor[HTML]{E0A4A4}} \color[HTML]{000000} 2140.4 & {\cellcolor[HTML]{E1E5BC}} \color[HTML]{000000} 491.7 & {\cellcolor[HTML]{B7D09D}} \color[HTML]{000000} 143.8 & {\cellcolor[HTML]{B0CD98}} \color[HTML]{000000} 81.2 & {\cellcolor[HTML]{A9C993}} \color[HTML]{000000} 21.8 & {\cellcolor[HTML]{A8C992}} \color[HTML]{000000} 11.4 \\
 &  & Francis & {\cellcolor[HTML]{E0A4A4}} \color[HTML]{000000} 1226.7 & {\cellcolor[HTML]{FDEFCE}} \color[HTML]{000000} 498.8 & {\cellcolor[HTML]{BED4A2}} \color[HTML]{000000} 120.0 & {\cellcolor[HTML]{B0CD98}} \color[HTML]{000000} 47.2 & {\cellcolor[HTML]{A9C993}} \color[HTML]{000000} 14.5 & {\cellcolor[HTML]{A8C992}} \color[HTML]{000000} 7.3 \\
 &  & Horse & {\cellcolor[HTML]{E0A4A4}} \color[HTML]{000000} 2604.0 & {\cellcolor[HTML]{D8E0B5}} \color[HTML]{000000} 497.3 & {\cellcolor[HTML]{B8D19E}} \color[HTML]{000000} 184.6 & {\cellcolor[HTML]{ADCB96}} \color[HTML]{000000} 68.5 & {\cellcolor[HTML]{A9C993}} \color[HTML]{000000} 22.6 & {\cellcolor[HTML]{A8C992}} \color[HTML]{000000} 10.8 \\
 &  & Lighthouse & {\cellcolor[HTML]{E0A4A4}} \color[HTML]{000000} 1324.7 & {\cellcolor[HTML]{FAF0CE}} \color[HTML]{000000} 476.6 & {\cellcolor[HTML]{DBE2B7}} \color[HTML]{000000} 270.6 & {\cellcolor[HTML]{AECC96}} \color[HTML]{000000} 40.8 & {\cellcolor[HTML]{A9C993}} \color[HTML]{000000} 14.7 & {\cellcolor[HTML]{A8C992}} \color[HTML]{000000} 7.2 \\
 &  & M60 & {\cellcolor[HTML]{E0A4A4}} \color[HTML]{000000} 1304.3 & {\cellcolor[HTML]{F9F0CD}} \color[HTML]{000000} 457.4 & {\cellcolor[HTML]{DBE2B7}} \color[HTML]{000000} 267.9 & {\cellcolor[HTML]{ADCB96}} \color[HTML]{000000} 34.6 & {\cellcolor[HTML]{A9C993}} \color[HTML]{000000} 14.3 & {\cellcolor[HTML]{A8C992}} \color[HTML]{000000} 7.2 \\
 &  & Panther & {\cellcolor[HTML]{E0A4A4}} \color[HTML]{000000} 2072.7 & {\cellcolor[HTML]{DFE4BA}} \color[HTML]{000000} 456.4 & {\cellcolor[HTML]{BCD3A1}} \color[HTML]{000000} 178.4 & {\cellcolor[HTML]{ADCB96}} \color[HTML]{000000} 52.0 & {\cellcolor[HTML]{A8C992}} \color[HTML]{000000} 13.9 & {\cellcolor[HTML]{A8C992}} \color[HTML]{000000} 7.2 \\
 &  & Playground & {\cellcolor[HTML]{E0A4A4}} \color[HTML]{000000} 1105.8 & {\cellcolor[HTML]{FEEDCD}} \color[HTML]{000000} 499.8 & {\cellcolor[HTML]{D5DFB2}} \color[HTML]{000000} 202.3 & {\cellcolor[HTML]{ACCB95}} \color[HTML]{000000} 25.6 & {\cellcolor[HTML]{AACA93}} \color[HTML]{000000} 17.8 & {\cellcolor[HTML]{A8C992}} \color[HTML]{000000} 8.4 \\
 &  & Train & {\cellcolor[HTML]{E0A4A4}} \color[HTML]{000000} 1097.9 & {\cellcolor[HTML]{FEEBCD}} \color[HTML]{000000} 526.7 & {\cellcolor[HTML]{D9E1B5}} \color[HTML]{000000} 217.0 & {\cellcolor[HTML]{AFCC97}} \color[HTML]{000000} 41.3 & {\cellcolor[HTML]{A9C993}} \color[HTML]{000000} 14.7 & {\cellcolor[HTML]{A8C992}} \color[HTML]{000000} 8.4 \\
\cmidrule{2-9}
 & \multirow[c]{7}{*}{\verttext{Train}} & Barn & {\cellcolor[HTML]{E0A4A4}} \color[HTML]{000000} 973.3 & {\cellcolor[HTML]{FEE8CC}} \color[HTML]{000000} 522.5 & {\cellcolor[HTML]{F9F0CD}} \color[HTML]{000000} 340.2 & {\cellcolor[HTML]{AECC96}} \color[HTML]{000000} 30.8 & {\cellcolor[HTML]{AACA93}} \color[HTML]{000000} 16.5 & {\cellcolor[HTML]{A8C992}} \color[HTML]{000000} 8.0 \\
 &  & Caterpillar & {\cellcolor[HTML]{E0A4A4}} \color[HTML]{000000} 1110.3 & {\cellcolor[HTML]{FEEBCD}} \color[HTML]{000000} 530.9 & {\cellcolor[HTML]{C9D9AA}} \color[HTML]{000000} 151.7 & {\cellcolor[HTML]{AECC96}} \color[HTML]{000000} 37.4 & {\cellcolor[HTML]{A9C993}} \color[HTML]{000000} 14.4 & {\cellcolor[HTML]{A8C992}} \color[HTML]{000000} 8.2 \\
 &  & Church & {\cellcolor[HTML]{E0A4A4}} \color[HTML]{000000} 1551.0 & {\cellcolor[HTML]{F7F0CC}} \color[HTML]{000000} 505.0 & {\cellcolor[HTML]{EBEAC3}} \color[HTML]{000000} 413.9 & {\cellcolor[HTML]{ACCB95}} \color[HTML]{000000} 38.5 & {\cellcolor[HTML]{A9C993}} \color[HTML]{000000} 19.1 & {\cellcolor[HTML]{A8C992}} \color[HTML]{000000} 8.4 \\
 &  & Courthouse & {\cellcolor[HTML]{E0A4A4}} \color[HTML]{000000} 1014.6 & {\cellcolor[HTML]{FEEACD}} \color[HTML]{000000} 508.3 & {\cellcolor[HTML]{F3EEC9}} \color[HTML]{000000} 305.9 & {\cellcolor[HTML]{B0CD98}} \color[HTML]{000000} 41.3 & {\cellcolor[HTML]{AACA93}} \color[HTML]{000000} 17.2 & {\cellcolor[HTML]{A8C992}} \color[HTML]{000000} 8.0 \\
 &  & Ignatius & {\cellcolor[HTML]{E0A4A4}} \color[HTML]{000000} 2724.5 & {\cellcolor[HTML]{D4DEB2}} \color[HTML]{000000} 485.2 & {\cellcolor[HTML]{B4CF9B}} \color[HTML]{000000} 144.7 & {\cellcolor[HTML]{ABCA94}} \color[HTML]{000000} 47.2 & {\cellcolor[HTML]{A8C992}} \color[HTML]{000000} 14.4 & {\cellcolor[HTML]{A8C992}} \color[HTML]{000000} 8.3 \\
 &  & Meetingroom & {\cellcolor[HTML]{E0A4A4}} \color[HTML]{000000} 1451.1 & {\cellcolor[HTML]{FDF0CF}} \color[HTML]{000000} 566.7 & {\cellcolor[HTML]{D3DEB1}} \color[HTML]{000000} 252.2 & {\cellcolor[HTML]{B3CE9A}} \color[HTML]{000000} 72.8 & {\cellcolor[HTML]{A9C993}} \color[HTML]{000000} 19.2 & {\cellcolor[HTML]{A8C992}} \color[HTML]{000000} 8.1 \\
 &  & Truck & {\cellcolor[HTML]{E0A4A4}} \color[HTML]{000000} 1549.9 & {\cellcolor[HTML]{F9F0CD}} \color[HTML]{000000} 535.1 & {\cellcolor[HTML]{C4D6A6}} \color[HTML]{000000} 183.0 & {\cellcolor[HTML]{B1CD98}} \color[HTML]{000000} 67.8 & {\cellcolor[HTML]{A8C992}} \color[HTML]{000000} 13.4 & {\cellcolor[HTML]{A8C992}} \color[HTML]{000000} 8.7 \\
\cmidrule{1-9} \cmidrule{2-9}
\multirow[c]{21}{*}{\verttext{100}} & \multirow[c]{6}{*}{\verttext{Advanced}} & Auditorium & {\cellcolor[HTML]{E0A4A4}} \color[HTML]{000000} 4885.4 & {\cellcolor[HTML]{D8E0B5}} \color[HTML]{000000} 931.7 & {\cellcolor[HTML]{CDDBAC}} \color[HTML]{000000} 740.7 & {\cellcolor[HTML]{ABCA94}} \color[HTML]{000000} 85.2 & {\cellcolor[HTML]{A8C992}} \color[HTML]{000000} 37.9 & {\cellcolor[HTML]{A8C992}} \color[HTML]{000000} 19.2 \\
 &  & Ballroom & {\cellcolor[HTML]{E0A4A4}} \color[HTML]{000000} 1987.4 & {\cellcolor[HTML]{FEEDCD}} \color[HTML]{000000} 909.4 & {\cellcolor[HTML]{FBF0CE}} \color[HTML]{000000} 732.2 & {\cellcolor[HTML]{E0E5BB}} \color[HTML]{000000} 448.7 & {\cellcolor[HTML]{A9C993}} \color[HTML]{000000} 26.5 & {\cellcolor[HTML]{A8C992}} \color[HTML]{000000} 17.5 \\
 &  & Courtroom & {\cellcolor[HTML]{E0A4A4}} \color[HTML]{000000} 4942.5 & {\cellcolor[HTML]{D7E0B4}} \color[HTML]{000000} 931.9 & {\cellcolor[HTML]{D3DEB1}} \color[HTML]{000000} 850.2 & {\cellcolor[HTML]{ADCB96}} \color[HTML]{000000} 127.7 & {\cellcolor[HTML]{A8C992}} \color[HTML]{000000} 28.9 & {\cellcolor[HTML]{A8C992}} \color[HTML]{000000} 17.0 \\
 &  & Museum & {\cellcolor[HTML]{E0A4A4}} \color[HTML]{000000} 5031.7 & {\cellcolor[HTML]{D2DDB0}} \color[HTML]{000000} 848.5 & {\cellcolor[HTML]{CEDBAD}} \color[HTML]{000000} 768.0 & {\cellcolor[HTML]{AFCC97}} \color[HTML]{000000} 172.2 & {\cellcolor[HTML]{A8C992}} \color[HTML]{000000} 34.6 & {\cellcolor[HTML]{A8C992}} \color[HTML]{000000} 16.5 \\
 &  & Palace & {\cellcolor[HTML]{ECBDB3}} \color[HTML]{000000} 1670.7 & {\cellcolor[HTML]{FEECCD}} \color[HTML]{000000} 885.1 & {\cellcolor[HTML]{E0A4A4}} \color[HTML]{000000} 1934.4 & {\cellcolor[HTML]{B1CD98}} \color[HTML]{000000} 87.4 & {\cellcolor[HTML]{AACA93}} \color[HTML]{000000} 32.2 & {\cellcolor[HTML]{A8C992}} \color[HTML]{000000} 15.7 \\
 &  & Temple & {\cellcolor[HTML]{E3AAA8}} \color[HTML]{000000} 1126.2 & {\cellcolor[HTML]{FBDAC5}} \color[HTML]{000000} 831.4 & {\cellcolor[HTML]{E0A4A4}} \color[HTML]{000000} 1169.3 & {\cellcolor[HTML]{B9D19F}} \color[HTML]{000000} 97.8 & {\cellcolor[HTML]{ABCA94}} \color[HTML]{000000} 31.9 & {\cellcolor[HTML]{A8C992}} \color[HTML]{000000} 16.5 \\
\cmidrule{2-9}
 & \multirow[c]{8}{*}{\verttext{Intermediate}} & Family & {\cellcolor[HTML]{E0A4A4}} \color[HTML]{000000} 2275.3 & {\cellcolor[HTML]{FDF0CF}} \color[HTML]{000000} 890.7 & {\cellcolor[HTML]{D1DDAF}} \color[HTML]{000000} 391.7 & {\cellcolor[HTML]{BFD4A3}} \color[HTML]{000000} 236.3 & {\cellcolor[HTML]{AACA93}} \color[HTML]{000000} 42.8 & {\cellcolor[HTML]{A8C992}} \color[HTML]{000000} 22.3 \\
 &  & Francis & {\cellcolor[HTML]{E0A4A4}} \color[HTML]{000000} 4689.7 & {\cellcolor[HTML]{D9E1B5}} \color[HTML]{000000} 909.8 & {\cellcolor[HTML]{BAD2A0}} \color[HTML]{000000} 365.1 & {\cellcolor[HTML]{AFCC97}} \color[HTML]{000000} 147.3 & {\cellcolor[HTML]{A8C992}} \color[HTML]{000000} 26.5 & {\cellcolor[HTML]{A8C992}} \color[HTML]{000000} 14.8 \\
 &  & Horse & {\cellcolor[HTML]{E0A4A4}} \color[HTML]{000000} 3882.0 & {\cellcolor[HTML]{E2E6BC}} \color[HTML]{000000} 909.5 & {\cellcolor[HTML]{BCD3A1}} \color[HTML]{000000} 351.9 & {\cellcolor[HTML]{B3CE9A}} \color[HTML]{000000} 189.2 & {\cellcolor[HTML]{A9C993}} \color[HTML]{000000} 43.5 & {\cellcolor[HTML]{A8C992}} \color[HTML]{000000} 21.1 \\
 &  & Lighthouse & {\cellcolor[HTML]{E0A4A4}} \color[HTML]{000000} 2590.9 & {\cellcolor[HTML]{F7F0CC}} \color[HTML]{000000} 845.2 & {\cellcolor[HTML]{D9E1B5}} \color[HTML]{000000} 512.1 & {\cellcolor[HTML]{B3CE9A}} \color[HTML]{000000} 128.9 & {\cellcolor[HTML]{A9C993}} \color[HTML]{000000} 28.5 & {\cellcolor[HTML]{A8C992}} \color[HTML]{000000} 14.0 \\
 &  & M60 & {\cellcolor[HTML]{E0A4A4}} \color[HTML]{000000} 1800.5 & {\cellcolor[HTML]{FEEDCD}} \color[HTML]{000000} 813.5 & {\cellcolor[HTML]{F0EDC6}} \color[HTML]{000000} 522.6 & {\cellcolor[HTML]{B7D09D}} \color[HTML]{000000} 127.2 & {\cellcolor[HTML]{A9C993}} \color[HTML]{000000} 25.9 & {\cellcolor[HTML]{A8C992}} \color[HTML]{000000} 14.5 \\
 &  & Panther & {\cellcolor[HTML]{E0A4A4}} \color[HTML]{000000} 1659.3 & {\cellcolor[HTML]{FEEBCD}} \color[HTML]{000000} 798.0 & {\cellcolor[HTML]{F1EDC7}} \color[HTML]{000000} 488.4 & {\cellcolor[HTML]{BFD4A3}} \color[HTML]{000000} 173.9 & {\cellcolor[HTML]{A9C993}} \color[HTML]{000000} 26.3 & {\cellcolor[HTML]{A8C992}} \color[HTML]{000000} 14.5 \\
 &  & Playground & {\cellcolor[HTML]{E0A4A4}} \color[HTML]{000000} 1303.4 & {\cellcolor[HTML]{FDDDC7}} \color[HTML]{000000} 888.5 & {\cellcolor[HTML]{F9F0CD}} \color[HTML]{000000} 456.9 & {\cellcolor[HTML]{B7D09D}} \color[HTML]{000000} 100.8 & {\cellcolor[HTML]{ABCA94}} \color[HTML]{000000} 33.0 & {\cellcolor[HTML]{A8C992}} \color[HTML]{000000} 17.0 \\
 &  & Train & {\cellcolor[HTML]{E0A4A4}} \color[HTML]{000000} 4441.7 & {\cellcolor[HTML]{DCE3B8}} \color[HTML]{000000} 930.8 & {\cellcolor[HTML]{D4DEB2}} \color[HTML]{000000} 789.5 & {\cellcolor[HTML]{AFCC97}} \color[HTML]{000000} 144.3 & {\cellcolor[HTML]{A8C992}} \color[HTML]{000000} 29.5 & {\cellcolor[HTML]{A8C992}} \color[HTML]{000000} 16.4 \\
\cmidrule{2-9}
 & \multirow[c]{7}{*}{\verttext{Train}} & Barn & {\cellcolor[HTML]{E0A4A4}} \color[HTML]{000000} 7502.6 & {\cellcolor[HTML]{C3D6A5}} \color[HTML]{000000} 811.7 & {\cellcolor[HTML]{BBD2A0}} \color[HTML]{000000} 605.9 & {\cellcolor[HTML]{AACA93}} \color[HTML]{000000} 98.0 & {\cellcolor[HTML]{A8C992}} \color[HTML]{000000} 30.7 & {\cellcolor[HTML]{A8C992}} \color[HTML]{000000} 16.6 \\
 &  & Caterpillar & {\cellcolor[HTML]{E0A4A4}} \color[HTML]{000000} 5145.0 & {\cellcolor[HTML]{D1DDAF}} \color[HTML]{000000} 857.4 & {\cellcolor[HTML]{B8D19E}} \color[HTML]{000000} 373.1 & {\cellcolor[HTML]{ADCB96}} \color[HTML]{000000} 120.9 & {\cellcolor[HTML]{A8C992}} \color[HTML]{000000} 25.8 & {\cellcolor[HTML]{A8C992}} \color[HTML]{000000} 16.8 \\
 &  & Church & {\cellcolor[HTML]{E0A4A4}} \color[HTML]{000000} 3242.8 & {\cellcolor[HTML]{E7E8C0}} \color[HTML]{000000} 821.2 & {\cellcolor[HTML]{E4E7BE}} \color[HTML]{000000} 782.0 & {\cellcolor[HTML]{B6CF9C}} \color[HTML]{000000} 201.9 & {\cellcolor[HTML]{A9C993}} \color[HTML]{000000} 34.5 & {\cellcolor[HTML]{A8C992}} \color[HTML]{000000} 16.6 \\
 &  & Courthouse & {\cellcolor[HTML]{E5ADA9}} \color[HTML]{000000} 990.9 & {\cellcolor[HTML]{F3C8BA}} \color[HTML]{000000} 838.3 & {\cellcolor[HTML]{E0A4A4}} \color[HTML]{000000} 1043.8 & {\cellcolor[HTML]{C3D6A5}} \color[HTML]{000000} 127.5 & {\cellcolor[HTML]{ACCB95}} \color[HTML]{000000} 32.0 & {\cellcolor[HTML]{A8C992}} \color[HTML]{000000} 15.2 \\
 &  & Ignatius & {\cellcolor[HTML]{E0A4A4}} \color[HTML]{000000} 2378.6 & {\cellcolor[HTML]{F7F0CB}} \color[HTML]{000000} 758.4 & {\cellcolor[HTML]{CFDCAE}} \color[HTML]{000000} 379.3 & {\cellcolor[HTML]{B7D09D}} \color[HTML]{000000} 156.1 & {\cellcolor[HTML]{A8C992}} \color[HTML]{000000} 25.4 & {\cellcolor[HTML]{A8C992}} \color[HTML]{000000} 16.3 \\
 &  & Meetingroom & {\cellcolor[HTML]{E0A4A4}} \color[HTML]{000000} 6338.4 & {\cellcolor[HTML]{CAD9AA}} \color[HTML]{000000} 874.5 & {\cellcolor[HTML]{BBD2A0}} \color[HTML]{000000} 526.6 & {\cellcolor[HTML]{AFCC97}} \color[HTML]{000000} 203.5 & {\cellcolor[HTML]{A8C992}} \color[HTML]{000000} 31.8 & {\cellcolor[HTML]{A8C992}} \color[HTML]{000000} 16.2 \\
 &  & Truck & {\cellcolor[HTML]{E0A4A4}} \color[HTML]{000000} 3367.2 & {\cellcolor[HTML]{E4E7BE}} \color[HTML]{000000} 805.5 & {\cellcolor[HTML]{CCDAAC}} \color[HTML]{000000} 495.9 & {\cellcolor[HTML]{B7D09D}} \color[HTML]{000000} 234.2 & {\cellcolor[HTML]{A8C992}} \color[HTML]{000000} 25.3 & {\cellcolor[HTML]{A8C992}} \color[HTML]{000000} 17.2 \\
\bottomrule
\end{tabular}

%% file: tables/supp/tt_runtimes_all_part2.tex
\begin{tabular}{lllcccccc}
\toprule
\multirow[c]{21}{*}{\verttext{200}} & \multirow[c]{6}{*}{\verttext{Advanced}} & Auditorium & {\cellcolor[HTML]{E0A4A4}} \color[HTML]{000000} 5388.1 & {\cellcolor[HTML]{F7F0CC}} \color[HTML]{000000} 1748.6 & {\cellcolor[HTML]{FDEFCF}} \color[HTML]{000000} 2185.9 & {\cellcolor[HTML]{B6CF9C}} \color[HTML]{000000} 349.0 & {\cellcolor[HTML]{A9C993}} \color[HTML]{000000} 71.1 & {\cellcolor[HTML]{A8C992}} \color[HTML]{000000} 37.1 \\
 &  & Ballroom & {\cellcolor[HTML]{E0A4A4}} \color[HTML]{000000} 3782.7 & {\cellcolor[HTML]{FEEDCE}} \color[HTML]{000000} 1680.6 & {\cellcolor[HTML]{FEECCD}} \color[HTML]{000000} 1779.0 & {\cellcolor[HTML]{FDF0CF}} \color[HTML]{000000} 1466.7 & {\cellcolor[HTML]{A9C993}} \color[HTML]{000000} 53.1 & {\cellcolor[HTML]{A8C992}} \color[HTML]{000000} 35.3 \\
 &  & Courtroom & {\cellcolor[HTML]{E0A4A4}} \color[HTML]{000000} 3245.0 & {\cellcolor[HTML]{FEE8CC}} \color[HTML]{000000} 1758.6 & {\cellcolor[HTML]{FEE6CB}} \color[HTML]{000000} 1886.8 & {\cellcolor[HTML]{C9D9AA}} \color[HTML]{000000} 453.8 & {\cellcolor[HTML]{A9C993}} \color[HTML]{000000} 58.1 & {\cellcolor[HTML]{A8C992}} \color[HTML]{000000} 34.6 \\
 &  & Museum & {\cellcolor[HTML]{E0A4A4}} \color[HTML]{000000} 3952.1 & {\cellcolor[HTML]{FDEECE}} \color[HTML]{000000} 1661.4 & {\cellcolor[HTML]{FEE8CC}} \color[HTML]{000000} 2162.8 & {\cellcolor[HTML]{CFDCAE}} \color[HTML]{000000} 634.0 & {\cellcolor[HTML]{AACA93}} \color[HTML]{000000} 68.7 & {\cellcolor[HTML]{A8C992}} \color[HTML]{000000} 33.2 \\
 &  & Palace & {\cellcolor[HTML]{EFC2B6}} \color[HTML]{000000} 3267.1 & {\cellcolor[HTML]{FEEDCD}} \color[HTML]{000000} 1747.5 & {\cellcolor[HTML]{E0A4A4}} \color[HTML]{000000} 3910.8 & {\cellcolor[HTML]{BAD2A0}} \color[HTML]{000000} 324.6 & {\cellcolor[HTML]{AACA93}} \color[HTML]{000000} 64.0 & {\cellcolor[HTML]{A8C992}} \color[HTML]{000000} 31.3 \\
 &  & Temple & {\cellcolor[HTML]{FDDFC8}} \color[HTML]{000000} 1480.1 & {\cellcolor[HTML]{F8D2C1}} \color[HTML]{000000} 1659.4 & {\cellcolor[HTML]{E0A4A4}} \color[HTML]{000000} 2221.5 & {\cellcolor[HTML]{C8D8A9}} \color[HTML]{000000} 309.2 & {\cellcolor[HTML]{ABCA94}} \color[HTML]{000000} 59.0 & {\cellcolor[HTML]{A8C992}} \color[HTML]{000000} 33.3 \\
\cmidrule{2-9}
 & \multirow[c]{8}{*}{\verttext{Intermediate}} & Family & {\cellcolor[HTML]{E0A4A4}} \color[HTML]{000000} 2349.0 & {\cellcolor[HTML]{FDDEC7}} \color[HTML]{000000} 1598.7 & {\cellcolor[HTML]{EEECC5}} \color[HTML]{000000} 679.2 & {\cellcolor[HTML]{EDEBC4}} \color[HTML]{000000} 667.0 & {\cellcolor[HTML]{ACCB95}} \color[HTML]{000000} 83.7 & {\cellcolor[HTML]{A8C992}} \color[HTML]{000000} 44.3 \\
 &  & Francis & {\cellcolor[HTML]{E0A4A4}} \color[HTML]{000000} 3522.3 & {\cellcolor[HTML]{FEECCD}} \color[HTML]{000000} 1654.0 & {\cellcolor[HTML]{E3E6BD}} \color[HTML]{000000} 846.6 & {\cellcolor[HTML]{C7D8A8}} \color[HTML]{000000} 462.9 & {\cellcolor[HTML]{A9C993}} \color[HTML]{000000} 55.8 & {\cellcolor[HTML]{A8C992}} \color[HTML]{000000} 29.6 \\
 &  & Horse & {\cellcolor[HTML]{E0A4A4}} \color[HTML]{000000} 4176.3 & {\cellcolor[HTML]{FDF0CF}} \color[HTML]{000000} 1610.1 & {\cellcolor[HTML]{CCDAAC}} \color[HTML]{000000} 624.4 & {\cellcolor[HTML]{C8D8A9}} \color[HTML]{000000} 558.8 & {\cellcolor[HTML]{AACA93}} \color[HTML]{000000} 88.9 & {\cellcolor[HTML]{A8C992}} \color[HTML]{000000} 41.9 \\
 &  & Lighthouse & {\cellcolor[HTML]{E0A4A4}} \color[HTML]{000000} 9072.2 & {\cellcolor[HTML]{D3DEB1}} \color[HTML]{000000} 1573.2 & {\cellcolor[HTML]{CFDCAE}} \color[HTML]{000000} 1411.4 & {\cellcolor[HTML]{B4CF9B}} \color[HTML]{000000} 455.9 & {\cellcolor[HTML]{A8C992}} \color[HTML]{000000} 54.5 & {\cellcolor[HTML]{A8C992}} \color[HTML]{000000} 28.0 \\
 &  & M60 & {\cellcolor[HTML]{E0A4A4}} \color[HTML]{000000} 5080.5 & {\cellcolor[HTML]{F1EDC7}} \color[HTML]{000000} 1481.3 & {\cellcolor[HTML]{E3E6BD}} \color[HTML]{000000} 1198.5 & {\cellcolor[HTML]{BCD3A1}} \color[HTML]{000000} 453.5 & {\cellcolor[HTML]{A9C993}} \color[HTML]{000000} 49.9 & {\cellcolor[HTML]{A8C992}} \color[HTML]{000000} 28.8 \\
 &  & Panther & {\cellcolor[HTML]{E0A4A4}} \color[HTML]{000000} 3837.3 & {\cellcolor[HTML]{FCF0CF}} \color[HTML]{000000} 1461.8 & {\cellcolor[HTML]{F2EEC8}} \color[HTML]{000000} 1134.0 & {\cellcolor[HTML]{CBDAAB}} \color[HTML]{000000} 560.2 & {\cellcolor[HTML]{A9C993}} \color[HTML]{000000} 49.8 & {\cellcolor[HTML]{A8C992}} \color[HTML]{000000} 28.7 \\
 &  & Playground & {\cellcolor[HTML]{E0A4A4}} \color[HTML]{000000} 8317.4 & {\cellcolor[HTML]{D7E0B4}} \color[HTML]{000000} 1578.9 & {\cellcolor[HTML]{C6D7A7}} \color[HTML]{000000} 1010.1 & {\cellcolor[HTML]{B2CE99}} \color[HTML]{000000} 360.0 & {\cellcolor[HTML]{A8C992}} \color[HTML]{000000} 61.4 & {\cellcolor[HTML]{A8C992}} \color[HTML]{000000} 33.1 \\
 &  & Train & {\cellcolor[HTML]{E0A4A4}} \color[HTML]{000000} 4022.3 & {\cellcolor[HTML]{FDF0CF}} \color[HTML]{000000} 1585.0 & {\cellcolor[HTML]{FDEFCE}} \color[HTML]{000000} 1638.5 & {\cellcolor[HTML]{C6D7A7}} \color[HTML]{000000} 503.0 & {\cellcolor[HTML]{A9C993}} \color[HTML]{000000} 54.3 & {\cellcolor[HTML]{A8C992}} \color[HTML]{000000} 32.9 \\
\cmidrule{2-9}
 & \multirow[c]{7}{*}{\verttext{Train}} & Barn & {\cellcolor[HTML]{E0A4A4}} \color[HTML]{000000} 6885.8 & {\cellcolor[HTML]{DEE4B9}} \color[HTML]{000000} 1504.8 & {\cellcolor[HTML]{FFFFFF}} \color[HTML]{000000} - & {\cellcolor[HTML]{B3CE9A}} \color[HTML]{000000} 349.2 & {\cellcolor[HTML]{A8C992}} \color[HTML]{000000} 56.6 & {\cellcolor[HTML]{A8C992}} \color[HTML]{000000} 33.5 \\
 &  & Caterpillar & {\cellcolor[HTML]{E0A4A4}} \color[HTML]{000000} 5360.8 & {\cellcolor[HTML]{F3EEC9}} \color[HTML]{000000} 1610.9 & {\cellcolor[HTML]{D9E1B5}} \color[HTML]{000000} 1066.4 & {\cellcolor[HTML]{B7D09D}} \color[HTML]{000000} 386.8 & {\cellcolor[HTML]{A8C992}} \color[HTML]{000000} 49.0 & {\cellcolor[HTML]{A8C992}} \color[HTML]{000000} 33.9 \\
 &  & Church & {\cellcolor[HTML]{E0A4A4}} \color[HTML]{000000} 4418.7 & {\cellcolor[HTML]{FAF0CD}} \color[HTML]{000000} 1577.1 & {\cellcolor[HTML]{FFFFFF}} \color[HTML]{000000} - & {\cellcolor[HTML]{C4D6A6}} \color[HTML]{000000} 522.5 & {\cellcolor[HTML]{AACA93}} \color[HTML]{000000} 69.7 & {\cellcolor[HTML]{A8C992}} \color[HTML]{000000} 34.3 \\
 &  & Courthouse & {\cellcolor[HTML]{E0A4A4}} \color[HTML]{000000} 8604.2 & {\cellcolor[HTML]{D6DFB3}} \color[HTML]{000000} 1598.8 & {\cellcolor[HTML]{FFFFFF}} \color[HTML]{000000} - & {\cellcolor[HTML]{B1CD98}} \color[HTML]{000000} 358.5 & {\cellcolor[HTML]{A8C992}} \color[HTML]{000000} 59.4 & {\cellcolor[HTML]{A8C992}} \color[HTML]{000000} 30.8 \\
 &  & Ignatius & {\cellcolor[HTML]{E0A4A4}} \color[HTML]{000000} 2565.2 & {\cellcolor[HTML]{FFE5CB}} \color[HTML]{000000} 1521.1 & {\cellcolor[HTML]{F8F0CC}} \color[HTML]{000000} 873.9 & {\cellcolor[HTML]{DAE1B6}} \color[HTML]{000000} 533.0 & {\cellcolor[HTML]{AACA93}} \color[HTML]{000000} 51.9 & {\cellcolor[HTML]{A8C992}} \color[HTML]{000000} 31.5 \\
 &  & Meetingroom & {\cellcolor[HTML]{E0A4A4}} \color[HTML]{000000} 4025.3 & {\cellcolor[HTML]{FDEFCE}} \color[HTML]{000000} 1645.4 & {\cellcolor[HTML]{FFFFFF}} \color[HTML]{000000} - & {\cellcolor[HTML]{D2DDB0}} \color[HTML]{000000} 697.6 & {\cellcolor[HTML]{A9C993}} \color[HTML]{000000} 59.6 & {\cellcolor[HTML]{A8C992}} \color[HTML]{000000} 32.0 \\
 &  & Truck & {\cellcolor[HTML]{E0A4A4}} \color[HTML]{000000} 3340.0 & {\cellcolor[HTML]{FEEDCD}} \color[HTML]{000000} 1531.7 & {\cellcolor[HTML]{F6F0CB}} \color[HTML]{000000} 1067.0 & {\cellcolor[HTML]{E8E9C1}} \color[HTML]{000000} 865.3 & {\cellcolor[HTML]{A9C993}} \color[HTML]{000000} 50.2 & {\cellcolor[HTML]{A8C992}} \color[HTML]{000000} 34.5 \\
\cmidrule{1-9} \cmidrule{2-9}
\multirow[c]{21}{*}{\verttext{full}} & \multirow[c]{6}{*}{\verttext{Advanced}} & Auditorium & {\cellcolor[HTML]{E0A4A4}} \color[HTML]{000000} 4713.9 & {\cellcolor[HTML]{FEEACD}} \color[HTML]{000000} 2349.4 & {\cellcolor[HTML]{FEE6CB}} \color[HTML]{000000} 2739.6 & {\cellcolor[HTML]{C3D6A5}} \color[HTML]{000000} 544.8 & {\cellcolor[HTML]{A9C993}} \color[HTML]{000000} 76.8 & {\cellcolor[HTML]{A8C992}} \color[HTML]{000000} 46.5 \\
 &  & Ballroom & {\cellcolor[HTML]{E0A4A4}} \color[HTML]{000000} 4462.5 & {\cellcolor[HTML]{FFE5CB}} \color[HTML]{000000} 2639.0 & {\cellcolor[HTML]{FFFFFF}} \color[HTML]{000000} - & {\cellcolor[HTML]{F7D0BF}} \color[HTML]{000000} 3407.7 & {\cellcolor[HTML]{A9C993}} \color[HTML]{000000} 83.6 & {\cellcolor[HTML]{A8C992}} \color[HTML]{000000} 55.7 \\
 &  & Courtroom & {\cellcolor[HTML]{E0A4A4}} \color[HTML]{000000} 3907.0 & {\cellcolor[HTML]{FEE0C8}} \color[HTML]{000000} 2556.5 & {\cellcolor[HTML]{FFFFFF}} \color[HTML]{000000} - & {\cellcolor[HTML]{E5E7BF}} \color[HTML]{000000} 975.6 & {\cellcolor[HTML]{A9C993}} \color[HTML]{000000} 80.5 & {\cellcolor[HTML]{A8C992}} \color[HTML]{000000} 50.4 \\
 &  & Museum & {\cellcolor[HTML]{E0A4A4}} \color[HTML]{000000} 5030.7 & {\cellcolor[HTML]{FEEBCD}} \color[HTML]{000000} 2422.9 & {\cellcolor[HTML]{FFFFFF}} \color[HTML]{000000} - & {\cellcolor[HTML]{E1E5BC}} \color[HTML]{000000} 1168.9 & {\cellcolor[HTML]{AACA93}} \color[HTML]{000000} 95.2 & {\cellcolor[HTML]{A8C992}} \color[HTML]{000000} 49.5 \\
 &  & Palace & {\cellcolor[HTML]{E0A4A4}} \color[HTML]{000000} 3856.3 & {\cellcolor[HTML]{E8B4AD}} \color[HTML]{000000} 3519.4 & {\cellcolor[HTML]{FFFFFF}} \color[HTML]{000000} - & {\cellcolor[HTML]{FDF0CF}} \color[HTML]{000000} 1537.4 & {\cellcolor[HTML]{ACCB95}} \color[HTML]{000000} 132.5 & {\cellcolor[HTML]{A8C992}} \color[HTML]{000000} 66.6 \\
 &  & Temple & {\cellcolor[HTML]{E0A4A4}} \color[HTML]{000000} 6103.9 & {\cellcolor[HTML]{FAF0CD}} \color[HTML]{000000} 2187.3 & {\cellcolor[HTML]{FFFFFF}} \color[HTML]{000000} - & {\cellcolor[HTML]{C6D7A7}} \color[HTML]{000000} 758.2 & {\cellcolor[HTML]{A9C993}} \color[HTML]{000000} 85.9 & {\cellcolor[HTML]{A8C992}} \color[HTML]{000000} 47.6 \\
\cmidrule{2-9}
 & \multirow[c]{8}{*}{\verttext{Intermediate}} & Family & {\cellcolor[HTML]{E1A5A5}} \color[HTML]{000000} 3655.3 & {\cellcolor[HTML]{E0A4A4}} \color[HTML]{000000} 3696.1 & {\cellcolor[HTML]{FFFFFF}} \color[HTML]{000000} - & {\cellcolor[HTML]{E7B3AD}} \color[HTML]{000000} 3396.7 & {\cellcolor[HTML]{AECC96}} \color[HTML]{000000} 207.5 & {\cellcolor[HTML]{A8C992}} \color[HTML]{000000} 110.1 \\
 &  & Francis & {\cellcolor[HTML]{E0A4A4}} \color[HTML]{000000} 4380.5 & {\cellcolor[HTML]{FEE8CC}} \color[HTML]{000000} 2369.8 & {\cellcolor[HTML]{FFFFFF}} \color[HTML]{000000} - & {\cellcolor[HTML]{DEE4B9}} \color[HTML]{000000} 961.9 & {\cellcolor[HTML]{AACA93}} \color[HTML]{000000} 80.0 & {\cellcolor[HTML]{A8C992}} \color[HTML]{000000} 45.2 \\
 &  & Horse & {\cellcolor[HTML]{E0A4A4}} \color[HTML]{000000} 4589.4 & {\cellcolor[HTML]{F4CBBC}} \color[HTML]{000000} 3641.7 & {\cellcolor[HTML]{FFFFFF}} \color[HTML]{000000} - & {\cellcolor[HTML]{FEE8CC}} \color[HTML]{000000} 2507.3 & {\cellcolor[HTML]{AFCC97}} \color[HTML]{000000} 223.4 & {\cellcolor[HTML]{A8C992}} \color[HTML]{000000} 100.4 \\
 &  & Lighthouse & {\cellcolor[HTML]{E0A4A4}} \color[HTML]{000000} 8594.3 & {\cellcolor[HTML]{E9E9C1}} \color[HTML]{000000} 2244.1 & {\cellcolor[HTML]{FFFFFF}} \color[HTML]{000000} - & {\cellcolor[HTML]{C2D5A5}} \color[HTML]{000000} 940.7 & {\cellcolor[HTML]{A9C993}} \color[HTML]{000000} 84.8 & {\cellcolor[HTML]{A8C992}} \color[HTML]{000000} 41.4 \\
 &  & M60 & {\cellcolor[HTML]{E0A4A4}} \color[HTML]{000000} 5796.9 & {\cellcolor[HTML]{FCF0CE}} \color[HTML]{000000} 2174.1 & {\cellcolor[HTML]{FFFFFF}} \color[HTML]{000000} - & {\cellcolor[HTML]{D6DFB3}} \color[HTML]{000000} 1094.1 & {\cellcolor[HTML]{A9C993}} \color[HTML]{000000} 80.2 & {\cellcolor[HTML]{A8C992}} \color[HTML]{000000} 44.4 \\
 &  & Panther & {\cellcolor[HTML]{E0A4A4}} \color[HTML]{000000} 4102.6 & {\cellcolor[HTML]{FEE9CC}} \color[HTML]{000000} 2174.4 & {\cellcolor[HTML]{FFFFFF}} \color[HTML]{000000} - & {\cellcolor[HTML]{F7F0CB}} \color[HTML]{000000} 1317.5 & {\cellcolor[HTML]{AACA93}} \color[HTML]{000000} 77.8 & {\cellcolor[HTML]{A8C992}} \color[HTML]{000000} 44.4 \\
 &  & Playground & {\cellcolor[HTML]{E0A4A4}} \color[HTML]{000000} 7335.7 & {\cellcolor[HTML]{F6F0CB}} \color[HTML]{000000} 2316.1 & {\cellcolor[HTML]{FFFFFF}} \color[HTML]{000000} - & {\cellcolor[HTML]{C5D7A7}} \color[HTML]{000000} 894.1 & {\cellcolor[HTML]{A9C993}} \color[HTML]{000000} 104.0 & {\cellcolor[HTML]{A8C992}} \color[HTML]{000000} 49.3 \\
 &  & Train & {\cellcolor[HTML]{E0A4A4}} \color[HTML]{000000} 7334.8 & {\cellcolor[HTML]{F8F0CC}} \color[HTML]{000000} 2487.8 & {\cellcolor[HTML]{FFFFFF}} \color[HTML]{000000} - & {\cellcolor[HTML]{CCDAAC}} \color[HTML]{000000} 1099.4 & {\cellcolor[HTML]{A9C993}} \color[HTML]{000000} 84.3 & {\cellcolor[HTML]{A8C992}} \color[HTML]{000000} 48.2 \\
\cmidrule{2-9}
 & \multirow[c]{7}{*}{\verttext{Train}} & Barn & {\cellcolor[HTML]{E0A4A4}} \color[HTML]{000000} 8072.1 & {\cellcolor[HTML]{FAF0CE}} \color[HTML]{000000} 2933.4 & {\cellcolor[HTML]{FFFFFF}} \color[HTML]{000000} - & {\cellcolor[HTML]{D7E0B4}} \color[HTML]{000000} 1558.5 & {\cellcolor[HTML]{A9C993}} \color[HTML]{000000} 123.1 & {\cellcolor[HTML]{A8C992}} \color[HTML]{000000} 66.1 \\
 &  & Caterpillar & {\cellcolor[HTML]{E0A4A4}} \color[HTML]{000000} 5909.6 & {\cellcolor[HTML]{FEEACC}} \color[HTML]{000000} 2989.8 & {\cellcolor[HTML]{FFFFFF}} \color[HTML]{000000} - & {\cellcolor[HTML]{D9E1B5}} \color[HTML]{000000} 1182.6 & {\cellcolor[HTML]{A9C993}} \color[HTML]{000000} 93.8 & {\cellcolor[HTML]{A8C992}} \color[HTML]{000000} 62.9 \\
 &  & Church & {\cellcolor[HTML]{E0A4A4}} \color[HTML]{000000} 5491.6 & {\cellcolor[HTML]{FEE1C8}} \color[HTML]{000000} 3590.2 & {\cellcolor[HTML]{FFFFFF}} \color[HTML]{000000} - & {\cellcolor[HTML]{FEE8CC}} \color[HTML]{000000} 2988.4 & {\cellcolor[HTML]{ACCB95}} \color[HTML]{000000} 175.1 & {\cellcolor[HTML]{A8C992}} \color[HTML]{000000} 84.4 \\
 &  & Courthouse & {\cellcolor[HTML]{E0A4A4}} \color[HTML]{000000} 11817.4 & {\cellcolor[HTML]{FFFFFF}} \color[HTML]{000000} - & {\cellcolor[HTML]{FFFFFF}} \color[HTML]{000000} - & {\cellcolor[HTML]{E9B6AE}} \color[HTML]{000000} 10656.7 & {\cellcolor[HTML]{ABCA94}} \color[HTML]{000000} 318.5 & {\cellcolor[HTML]{A8C992}} \color[HTML]{000000} 177.4 \\
 &  & Ignatius & {\cellcolor[HTML]{E0A4A4}} \color[HTML]{000000} 2407.4 & {\cellcolor[HTML]{FFFFFF}} \color[HTML]{000000} - & {\cellcolor[HTML]{FFFFFF}} \color[HTML]{000000} - & {\cellcolor[HTML]{F9F0CD}} \color[HTML]{000000} 844.2 & {\cellcolor[HTML]{AACA93}} \color[HTML]{000000} 67.1 & {\cellcolor[HTML]{A8C992}} \color[HTML]{000000} 40.6 \\
 &  & Meetingroom & {\cellcolor[HTML]{E0A4A4}} \color[HTML]{000000} 4440.0 & {\cellcolor[HTML]{FFFFFF}} \color[HTML]{000000} - & {\cellcolor[HTML]{FFFFFF}} \color[HTML]{000000} - & {\cellcolor[HTML]{FEE8CC}} \color[HTML]{000000} 2422.2 & {\cellcolor[HTML]{AACA93}} \color[HTML]{000000} 103.9 & {\cellcolor[HTML]{A8C992}} \color[HTML]{000000} 58.7 \\
 &  & Truck & {\cellcolor[HTML]{E0A4A4}} \color[HTML]{000000} 3488.5 & {\cellcolor[HTML]{FFFFFF}} \color[HTML]{000000} - & {\cellcolor[HTML]{FEEECE}} \color[HTML]{000000} 1528.8 & {\cellcolor[HTML]{FAF0CE}} \color[HTML]{000000} 1274.3 & {\cellcolor[HTML]{A9C993}} \color[HTML]{000000} 63.3 & {\cellcolor[HTML]{A8C992}} \color[HTML]{000000} 41.5 \\
 \bottomrule
\end{tabular}